\documentclass[onecolumn]{c2alab}           %

\metadata[Website]{\url{https://hex-humanoid.github.io/}}
\metadata[Github]{\url{https://github.com/Open-X-Humanoid/HEX}}

\usepackage{tikz}
\usetikzlibrary{tikzmark, calc}
\usepackage{makecell}

\usepackage{svg} 
\usepackage{wrapfig}
\usepackage{makecell}

\title{HEX: Humanoid-Aligned Experts for Cross-Embodiment Whole-Body Manipulation}

\author{
\begin{center}
    Shuanghao Bai$^{1,2*}$ \quad
    Meng Li$^{1*\dagger}$ \quad
    Xinyuan Lv$^{1,3}$ \quad
    Jiawei Wang$^{1,3}$ \quad
    Xinhua Wang$^{1}$ \quad
    Fei Liao$^{1}$ \\ 
    Chengkai Hou$^{1,4}$ \quad
    Langzhe Gu$^{1,4}$ \quad
    Bowei Zhang$^{1,3}$ \quad
    Wanqi Zhou$^{2}$ \quad
    Kun Wu$^{1}$ \quad
    Ziluo Ding$^{1}$ \\ 
    Zhiyuan Xu$^{1}$ \quad
    Lei Sun$^{3}$ \quad
    Shanghang Zhang$^{4}$ \quad
    Zhengping Che$^{1\ddagger}$ \quad
    Jian Tang$^{1\ddagger}$ \quad
    Badong Chen$^{2\ddagger}$
    \\[10pt]
    {\small
    $^{1}$Beijing Innovation Center of Humanoid Robotics \quad
    $^{2}$Xi'an Jiaotong University \\
    $^{3}$Nankai University \quad
    $^{4}$Peking University
    \\[8pt]
    $^{*}$Equal Contribution \qquad
    $^{\dagger}$Project Lead \qquad
    $^{\ddagger}$Corresponding Author}
\end{center}
}

\begin{document}

\abstract{
Humans achieve complex manipulation through coordinated whole-body control, whereas most Vision-Language-Action (VLA) models treat robot body parts largely independently, making high-DoF humanoid control challenging and often unstable. We present HEX, a state-centric framework for coordinated manipulation on full-sized bipedal humanoid robots. HEX introduces a humanoid-aligned universal state representation for scalable learning across heterogeneous embodiments, and incorporates a Mixture-of-Experts Unified Proprioceptive Predictor to model whole-body coordination and temporal motion dynamics from large-scale multi-embodiment trajectory data. To efficiently capture temporal visual context, HEX uses lightweight history tokens to summarize past observations, avoiding repeated encoding of historical images during inference. It further employs a residual-gated fusion mechanism with a flow-matching action head to adaptively integrate visual-language cues with proprioceptive dynamics for action generation. Experiments on real-world humanoid manipulation tasks show that HEX achieves state-of-the-art performance in task success rate and generalization, particularly in fast-reaction and long-horizon scenarios.
}

\maketitle

\justifying
\section{Introduction}
\label{sec:introduction}

\begin{figure}[!t]
\centering
\includegraphics[width=0.95\textwidth]{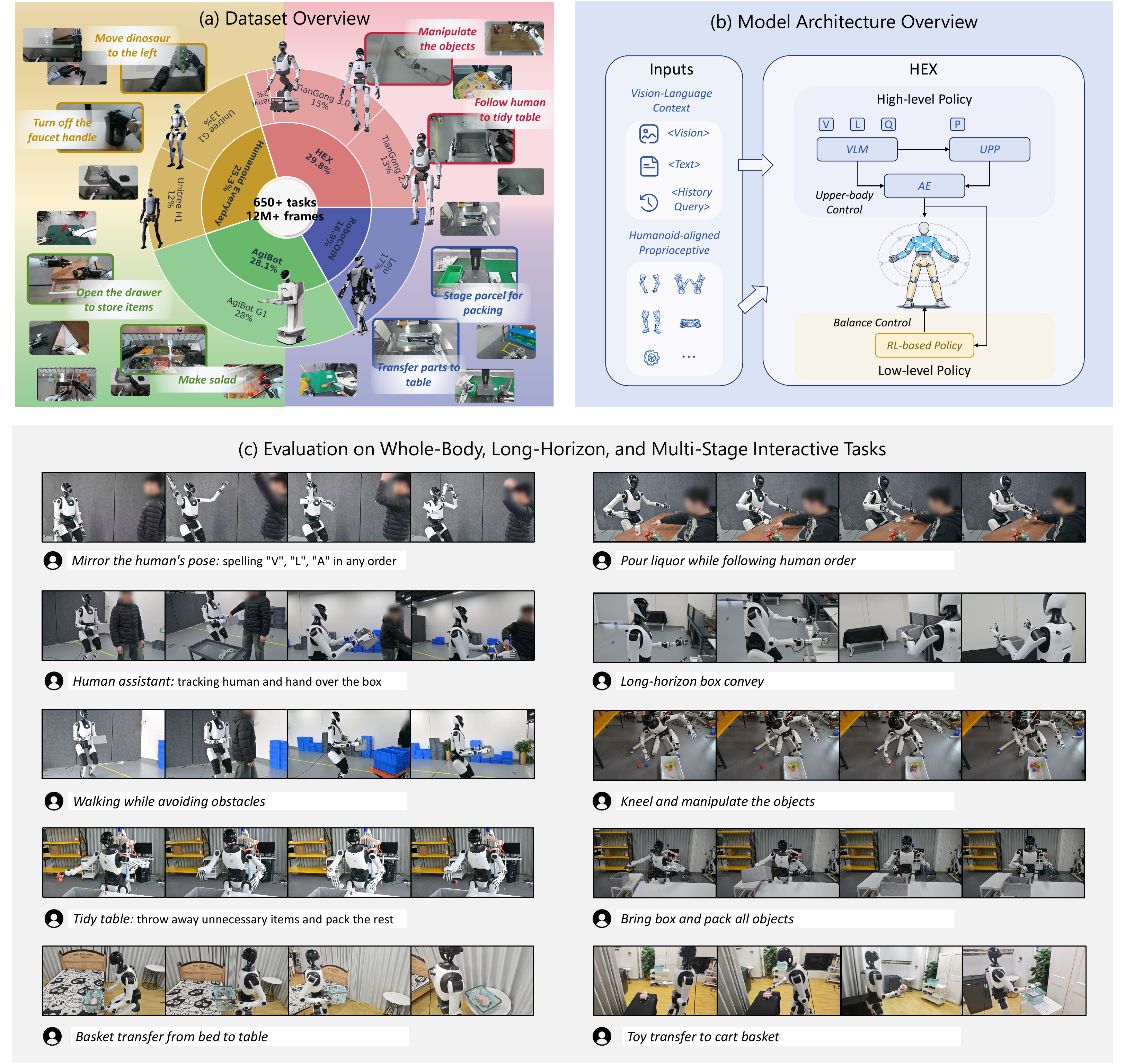}
\vskip -0.08in
\caption{
\textbf{Overview of HEX.}
    (a) HEX is, to the best of our knowledge, the first whole-body VLA framework for full-sized bipedal humanoid robots, pretrained on diverse cross-embodiment humanoid trajectory data.
    (b) HEX combines a high-level VLA module with a low-level whole-body controller for coordinated action generation and balance-preserving execution.
    (c) We evaluate HEX on TianGong 2.0 and TianGong 3.0 across whole-body, long-horizon, and fast-reaction tasks, demonstrating strong performance across diverse manipulation scenarios.
}
\label{fig:teaser}
\end{figure}

Humanoid robots hold the promise of bringing embodied intelligence into complex human environments such as homes and schools. Existing research, however, has largely focused on either locomotion, which enables robots to navigate unstructured environments~\cite{nakanishi2004learning,peng2018deepmimic,radosavovic2024real}, or hand-centric manipulation, where the lower body remains fixed and control is limited to the arms and hands~\cite{li2025okami,yang2025egovla}. In contrast, humans routinely perform tasks that require simultaneous locomotion and manipulation, leveraging coordinated motion of the entire body. Enabling such whole-body manipulation in humanoids remains significantly underexplored. The challenge is fundamental: the robot must maintain dynamic balance while producing high-dimensional, tightly coupled motions across multiple limbs during object interaction.

Existing approaches to humanoid whole-body manipulation mainly follow two paradigms. The first adopts an explicitly decomposed design, in which locomotion or navigation and manipulation are controlled by separate policies~\cite{xie2023hierarchical}. While such decomposition simplifies learning and control, it relies heavily on manual task priors and interface design, and becomes increasingly brittle as task complexity grows. Errors can also accumulate across modules, making tightly coupled behaviors, such as manipulation during locomotion, difficult to achieve robustly. A more recent trend is to adopt a hierarchical design, where the high-level module produces task-relevant commands, such as arm and hand actions~\cite{liu2026trajbooster, jiang2026wholebodyvla} or corresponding hand-eye targets~\cite{chen2025hand}, while a low-level whole-body controller refines them into high-frequency, balance-preserving motions.

In parallel, recent Vision-Language-Action (VLA) models have introduced stronger visual-semantic understanding and reasoning through large vision-language models, showing promising scalability and generalization~\cite{kim2025openvla, intelligence2025pi05, cui2025openhelix, fan2025long}. As a result, recent humanoid systems increasingly adopt VLA-style high-level planners together with low-level whole-body controllers for stable execution~\cite{liu2026trajbooster, jiang2026wholebodyvla, bjorck2025gr00t, wei2026psi0}. Despite these advances, most existing VLA-based approaches remain insufficiently structured for humanoid whole-body manipulation. In many cases, actions are predicted over high-dimensional joints or latent commands without explicitly modeling how body parts interact through shared balance and posture. As a result, the policy may capture task intent semantically, yet still fail to produce coordinated whole-body behavior, especially in fast-reaction and long-horizon scenarios where temporal consistency and whole-body coordination are essential.

To address this limitation, we propose \textbf{HEX}, a framework built on the key insight that effective humanoid whole-body manipulation requires both embodiment-aware predictive dynamics and temporally grounded scene understanding, as shown in Figure~\ref{fig:teaser}. Specifically, HEX introduces a humanoid-aligned universal state representation that provides a structured basis for modeling whole-body proprioceptive dynamics across different body parts and embodiments. Built upon this representation, HEX captures temporal motion evolution and whole-body coordination in proprioceptive space, enabling scalable predictive modeling for heterogeneous humanoid trajectories.
Whole-body manipulation also depends on temporal visual context, especially when object motion, scene evolution, or partial observability makes the current observation insufficient. To this end, HEX summarizes past visual-language context into compact representations while leveraging predictive proprioceptive dynamics to provide state foresight. Together, these designs form a \textit{review-and-forecast} paradigm, where past visual context supports scene understanding and future state prediction supports coordinated whole-body control. The resulting visual-language and predictive state representations are then adaptively fused for action generation, producing smooth and coordinated whole-body behaviors.

We evaluate HEX on a diverse set of real-world humanoid manipulation tasks against strong VLA and imitation learning baselines, including ACT~\cite{zhao2023learning}, SwitchVLA~\cite{li2025switchvla}, GR00T N1.5~\cite{bjorck2025gr00t}, $\Pi_{0.5}$~\cite{intelligence2025pi05}, and $\psi_0$~\cite{wei2026psi0}. Across a wide range of task settings, HEX consistently achieves higher task success rates and stronger generalization. The improvements are particularly pronounced in fast-reaction and long-horizon scenarios, where coordinated whole-body dynamics and temporal consistency are critical.
In summary, our contributions are fourfold. First, to the best of our knowledge, we present the first whole-body VLA framework for full-sized bipedal humanoid robots. Second, we propose a cross-embodiment humanoid-aligned state representation with predictive proprioceptive modeling for scalable whole-body pretraining. Third, we introduce a review-and-forecast paradigm that combines visual history summarization, future state prediction, and adaptive multimodal fusion for action generation. Finally, extensive experiments on real-world humanoid manipulation benchmarks demonstrate state-of-the-art performance and validate HEX as an effective framework for coordinated whole-body manipulation.

\section{Related Work}
\label{sec:related_work}

\subsection{Learning-based Humanoid Whole-body Control}

Learning-based humanoid whole-body control has been primarily advanced through reinforcement learning (RL) and imitation learning (IL)~\cite{bai2025towards, ze2025twist, li2026language, li2026robomirror}.
Early RL-based methods such as DeepMimic~\cite{peng2018deepmimic} and AMP~\cite{peng2021amp} established motion-tracking and motion-prior-based policy learning as effective paradigms for acquiring robust and natural humanoid skills. More recent work has extended this line toward real-world, contact-rich, and visually grounded whole-body control, including simulation-pretrained latent action learning for real-world RL~\cite{hu2025slac}, force-adaptive loco-manipulation~\cite{zhang2026falcon}, highly dynamic full-body skill learning~\cite{xie2026kungfubot}, large-scale motion-tracking controllers with strong generalization~\cite{luo2026sonic}, and visual sim-to-real humanoid loco-manipulation via privileged RL and teacher--student policy distillation~\cite{he2026viral}. In parallel, imitation learning has emerged as an efficient alternative by leveraging human demonstrations, teleoperation trajectories, and motion priors. Recent approaches explore human-to-humanoid imitation from teleoperation or egocentric demonstrations~\cite{fu2025humanplus,he2025omnih2o,qiu2025humanoid}, unified motion-tracking and predictive motion priors~\cite{chen2025gmt,lu2025mobile}, as well as generative imitation with diffusion-based policies~\cite{liao2026beyondmimic,ze2025generalizable}. Despite their success, these methods are primarily designed for skill imitation or task-specific control, and generally provide limited semantic understanding of instructions, goals, and visual context.

More recently, vision-language-action (VLA) models, empowered by the strong visual-semantic understanding and reasoning capabilities of large vision--language models and their potential to scale to more general scenarios~\cite{bai2026embodied,bu2025univla,song2026reconvla,wang2026vlaadapter,bai2026latent}, have begun to extend from fixed-base manipulation to humanoid whole-body control. Humanoid-VLA~\cite{ding2025humanoid} introduces visual integration for humanoid control, while GR00T N1~\cite{bjorck2025gr00t} and $\Psi_0$~\cite{wei2026psi0} move toward generalist humanoid foundation models trained on large-scale heterogeneous data. To better handle whole-body behaviors, LeVERB~\cite{xue2025leverb} proposes hierarchical latent vision-language instructions, WholeBodyVLA~\cite{jiang2026wholebodyvla} explores unified latent VLA control for loco-manipulation, and TrajBooster~\cite{liu2026trajbooster} improves downstream adaptation through trajectory-centric retargeting. More recent approaches explore complementary strategies: OpenHLM~\cite{hu2026openhlm} investigates whole-body teleoperation and heterogeneous co-training, VLK~\cite{wang2026vlk} leverages synthetic interactions in reconstructed scenes, HALOMI~\cite{zhao2026halomi} incorporates active perception and human demonstrations, and HAF~\cite{gu2026haf} combines hierarchical action flow with latent reinforcement learning.
In parallel, world-action models (WAMs) have emerged as an alternative paradigm for coupling predictive dynamics with whole-body action generation. MotionWAM~\cite{zheng2026motionwam} exploits video-model features and unified motion representations for real-time loco-manipulation, while $\omega$-0~\cite{li2026omega} integrates latent future-observation prediction with whole-body action generation. Two recent WholeBodyWAM approaches further investigate scalable motion priors~\cite{zhang2026wholebodywam} and pretrained world-action priors with whole-body-controller-grounded coordination~\cite{li2026wholebodywam}, respectively. 

Complementary to these advances in data-driven adaptation, action generation, and predictive modeling, our work focuses on explicitly capturing structured proprioceptive dependencies through humanoid-aligned state representations and joint past-future temporal modeling, coupling visual-language reasoning with embodiment-aware representations for coordinated whole-body control.

\subsection{Cross-Embodiment Learning for Humanoid Robots}

Cross-embodiment learning seeks to transfer knowledge across agents with different morphologies by learning shared behavior representations, aligned control spaces, or generalizable pretrained policies~\cite{wang2024scaling,yang2024pushing,doshi2025scaling}. 
One line of work focuses on \emph{human-to-humanoid learning}, where human videos or egocentric demonstrations are used as scalable supervision for humanoid control~\cite{mao2025universal,qiu2025humanoid,weng2025hdmi,shi2026egohumanoid,yang2026zerowbc,wei2026psi0}. Representative examples include Mao et al.~\cite{mao2025universal}, which leverage large-scale human videos for humanoid pose control, Humanoid Policy$\sim$Human Policy~\cite{qiu2025humanoid}, which aligns egocentric human demonstrations with humanoid behaviors in a unified policy space, and $\Psi_0$~\cite{wei2026psi0}, which incorporates human egocentric videos into a staged humanoid foundation-model training recipe.
Another line studies \emph{robot-to-humanoid} or \emph{cross-humanoid learning}, where policies or representations are transferred across heterogeneous robotic embodiments~\cite{lin2026h,punamiya2025egobridge,yang2025egovla,xue2026scalable,peng2026embodiment,bjorck2025gr00t,luo2026being}. Representative works include H-Zero~\cite{lin2026h}, which enables few-shot transfer to novel humanoids through cross-humanoid pretraining, EAGLE~\cite{peng2026embodiment}, which learns a unified controller across diverse humanoid embodiments, and GR00T N1~\cite{bjorck2025gr00t}, which scales humanoid foundation modeling with heterogeneous robot trajectories, human videos, and synthetic data. 

Similar in spirit, HEX also targets cross-embodiment humanoid learning, but differs from prior works by introducing a compositional and humanoid-aligned proprioceptive modeling framework. Its Unified Proprioceptive Predictor with Mixture-of-Experts~(MoE) operates on canonical body-part abstractions, allowing heterogeneous trajectories from the same embodiment or different humanoids to be encoded in a shared latent space without retraining a monolithic state encoder for every new joint configuration or missing-part setting. By combining reusable part-level encoders with dynamic expert routing, HEX more efficiently exploits both intra- and cross-embodiment data, while capturing structured whole-body and temporal dependencies for coordinated whole-body control.
\section{Method}
\label{sec:method}

\subsection{Overview}
\label{subsec:overview}

HEX adopts a hierarchical architecture for humanoid whole-body manipulation, consisting of a high-level VLA policy and a low-level RL-based whole-body controller. The high-level policy takes visual-language context together with humanoid-aligned proprioceptive states as input and produces task-conditioned outputs. Specifically, it directly predicts upper-body manipulation actions for the arms and hands, while simultaneously generating lower-body commands for the whole-body controller. These commands specify both the motion mode, such as standing, walking, or kneeling, and the desired motion within each mode. The low-level controller operates at a higher control frequency and translates these commands into dynamically stable joint-level motions while maintaining balance through state feedback.

\begin{figure*}[t]
\centering
\includegraphics[width=\textwidth]{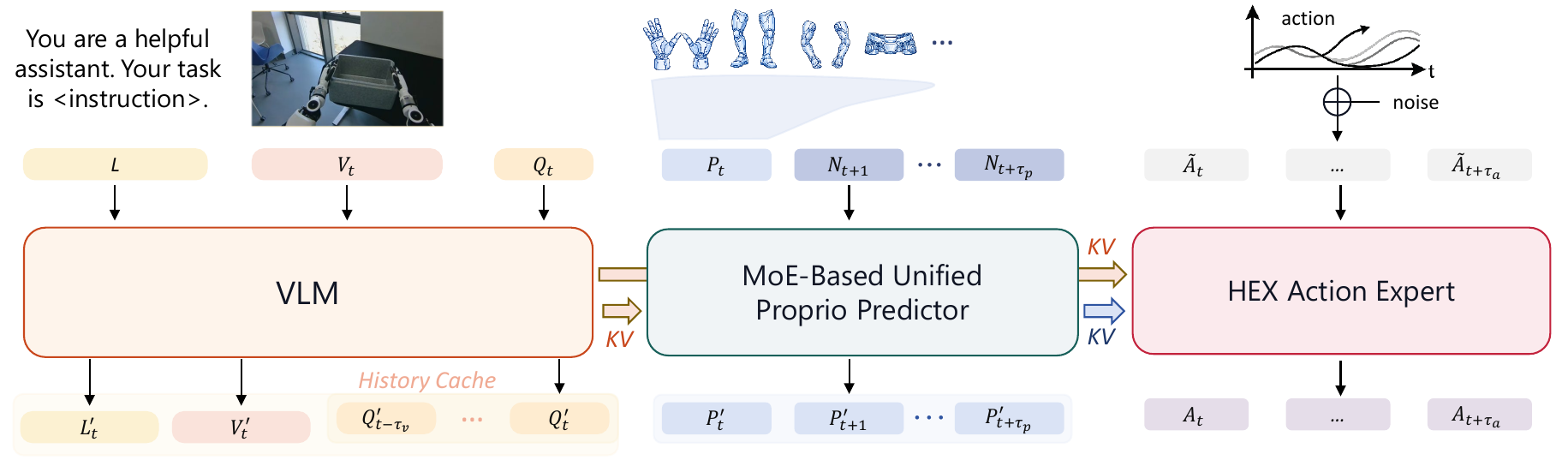}
\vskip -0.08in
\caption{
\textbf{Overview of the proposed high-level VLA policy in HEX.}
Given a language instruction $L$, the current visual observation $V_t$, and a history query token $Q_t$, the VLM encodes visual-language context together with lightweight temporal review cues summarized in a history cache. In parallel, humanoid-aligned proprioceptive states are organized into structured part-aware tokens and processed by a MoE-based Unified Proprioceptive Predictor, which captures whole-body interactions and forecasts future state dynamics. The resulting visual-language and predictive proprioceptive features are then integrated by the HEX Action Expert through adaptive fusion for action generation, producing task-relevant high-level actions over the prediction horizon.
}
\label{fig:model}
\end{figure*}

The high-level VLA policy in HEX consists of three main components. First, a Visual-Language Model (VLM) module encodes current visual-language context together with lightweight temporal review cues. Second, a Unified Proprioceptive Predictor (UPP) models humanoid-aligned state dynamics and captures whole-body interactions through predictive proprioceptive modeling. Third, an action expert integrates visual-language and proprioceptive features through adaptive fusion to generate the final high-level action. Figure~\ref{fig:model} provides an overview of the full framework.
For the low-level controller, we instantiate skill-specific RL policies trained with motion-guided objectives. These policies share a unified command interface, allowing the high-level VLA to select the appropriate motion regime and specify the desired lower-body motion through the same control interface. In particular, the standing and walking controllers are trained with a DeepMimic-style reference-tracking formulation~\cite{peng2018deepmimic}, which is well suited to stable periodic or quasi-static motions with clear target kinematics. In contrast, the half-kneeling controller is trained with an AMP-style objective~\cite{peng2021amp}, where an adversarial motion prior encourages natural contact-rich posture transitions without requiring strict frame-wise tracking to a single reference trajectory.
In the following, we present the core components of the high-level VLA policy in HEX.

\subsection{VLM with History Query Feature Cache}

To incorporate temporal visual-language context without repeatedly encoding long image histories, we introduce a lightweight history query feature cache. At each timestep $t$, we encode the language instruction, the current visual observation, and a query token using a single vision--language model (VLM). Specifically, we concatenate the language tokens $\mathbf{L}$, visual tokens $\mathbf{V}_t$ extracted from observation $\mathbf{o}_t$, and a query token $\mathbf{Q}_t$, and feed them into the VLM:
\begin{equation}
[\mathbf{L}'_t, \mathbf{V}'_t, \mathbf{Q}'_t]
=
f_{\mathrm{vlm}}([\mathbf{L}, \mathbf{V}_t, \mathbf{Q}_t]).
\end{equation}
The resulting query feature $\mathbf{Q}'_t$ serves as a compact summary of the current visual-language context.
Rather than propagating the query token itself across time, we generate a fresh query feature at every timestep and store it in a fixed-length cache:
\begin{equation}
\mathcal{M}^{\mathrm{vl}}_t
=
\{\mathbf{Q}'_{t-\tau_v}, \dots, \mathbf{Q}'_{t-1}, \mathbf{Q}'_t\},
\end{equation}
where $\tau_v$ denotes the visual history window, set to 2 in all experiments. This cache stores only compact query features, rather than the original images or full VLM activations. Together with the current-step visual-language features $\mathbf{L}'_t$ and $\mathbf{V}'_t$, $\mathcal{M}^{\mathrm{vl}}_t$ provides recent semantic context for the subsequent proprioceptive modeling and action generation modules.

This design provides an efficient form of visual review: temporal scene information is preserved through a compact feature memory, while the VLM itself remains a single-step feed-forward encoder applied only to the current frame. As a result, HEX can exploit temporal visual context without incurring the substantial cost of repeatedly encoding long visual histories.

\subsection{UPP with Morphology-based MoE}
\label{subsec:state}

Humanoid proprioceptive states comprise heterogeneous signals, including whole-body joint positions, velocities, accelerations, and tactile feedback. As sensing capabilities expand, policies must capture both these signals and their interactions across body parts to support coordinated whole-body control.

To enable cross-embodiment learning, we organize proprioceptive states into canonical body-part slots, including left/right arms, left/right hands, left/right legs, head, waist, and an auxiliary \texttt{others} slot.
This decomposition follows two considerations. First, it preserves body-part-specific motion characteristics before cross-part interactions are modeled by UPP. Second, the fine-grained partition facilitates cross-embodiment adaptation by localizing embodiment-specific variations. For example, humanoid platforms may share similar arm structures while employing substantially different end-effectors, such as dexterous hands or parallel grippers. Separating arms from hands therefore allows transferable arm representations to be preserved when only the end-effector changes, whereas merging them would entangle these variations within a single representation.
Although we use this set of canonical slots in the current work, the formulation is readily extensible to richer or more fine-grained body-part decompositions.

For an embodiment $e$, the raw proprioceptive state $\mathbf{s}^{(e)}_t$ may vary in dimensionality and composition.
We first partition $\mathbf{s}^{(e)}_t$ into the available body-part observations. Each body part is then processed by a lightweight proprioceptive encoder and projected into the same $d$-dimensional latent space. For unavailable body parts, the latent is replaced by a learned missing-part token. The resulting body-part latent tokens are concatenated to form the structured representation
\begin{equation}
\mathbf{P}^{(e)}_t \in \mathbb{R}^{P \times d},
\end{equation}
where $P$ denotes the number of canonical part slots and $d$ is the latent dimension. This enables prediction in a shared latent space rather than the raw embodiment-specific state space.

\begin{figure*}[t]
\centering
\includegraphics[width=0.95\textwidth]{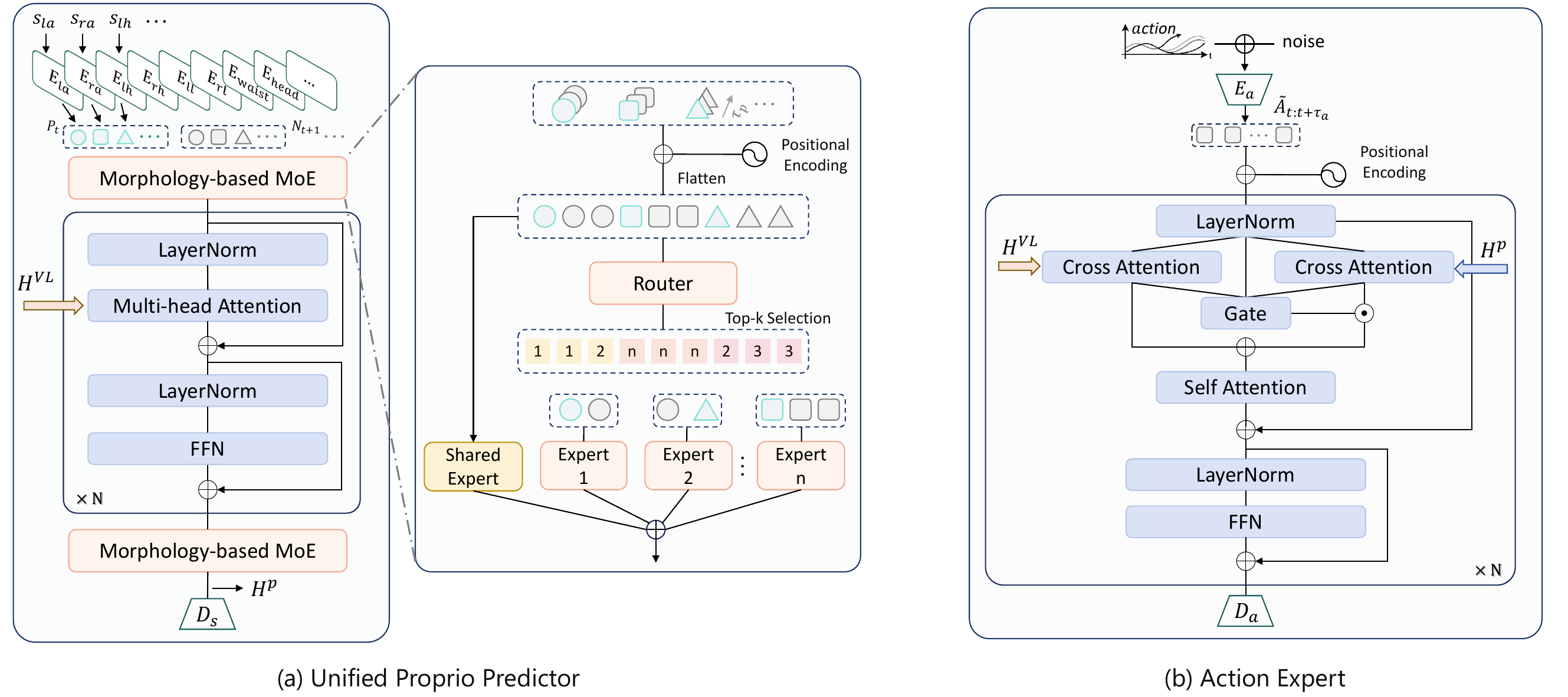}
\vskip -0.08in
\caption{
\textbf{Left and middle: Unified Proprioceptive Predictor (UPP).} Morphology-based proprioceptive states are first mapped into canonical body-part tokens and augmented with learnable future query tokens. These spatio-temporal tokens are processed by a shared transformer backbone sandwiched by morphology-aware MoE adaptation modules, yielding future proprioceptive latents $\mathbf{H}^{p}$. The middle panel details the morphology-aware MoE, where flattened part-time tokens are routed by a learned top-$k$ gate to a set of routed experts, while a shared expert provides a common transformation across all tokens. This design enables token-wise specialization for embodiment- and part-dependent variations while preserving reusable dynamics across embodiments.
\textbf{Right: Action Expert.} Noisy action tokens are encoded and conditioned on both visual-language features $\mathbf{H}^{VL}$ and predicted proprioceptive features $\mathbf{H}^{p}$ through dual cross-attention. A learned gate adaptively injects the state branch on top of the visual-language branch, followed by self-attention and feed-forward refinement. The resulting denoised features are decoded into high-level actions for arm and hand control and for downstream whole-body execution.
}
\label{fig:upp_model}
\end{figure*}

While part-aligned representations accommodate heterogeneous states, whole-body prediction also requires modeling temporal evolution and cross-part interactions. HEX therefore employs a Unified Proprioceptive Predictor (UPP), illustrated in Figure~\ref{fig:upp_model}(a), to capture these dependencies. We concatenate the current part tokens with learnable future query tokens $\mathbf{N}_{t+1:t+\tau_p}$, where $\tau_p$ denotes the prediction horizon, and add temporal and part positional embeddings to form a spatio-temporal token sequence.

To better accommodate embodiment- and token-specific variations while preserving a shared predictive backbone, UPP incorporates two lightweight morphology-aware MoE modules: an input-side MoE before the shared transformer predictor and an output-side MoE after it.
As shown in Figure~\ref{fig:upp_model}(a), each flattened spatio-temporal token is routed by a learned top-$k$ gate to selected experts, alongside a shared expert branch. The routed expert outputs are weighted and combined with the shared expert output. This enables specialization in embodiment- and part-specific dynamics while preserving reusable knowledge across embodiments. The MoE modules thus provide lightweight token-level adaptation around the shared transformer backbone.

Between the input-side and output-side MoE adaptation modules, a shared transformer backbone models embodiment-agnostic temporal dynamics in the latent space.
The backbone processes the part-time tokens using interleaved self-attention and visual-language-conditioned attention to capture cross-part dependencies and task-relevant dynamics. Conditioned on the current proprioceptive latent $\mathbf{P}_t$, language and visual features $\mathbf{L}'_t$ and $\mathbf{V}'_t$, and visual-language history cache $\mathcal{M}^{\mathrm{vl}}_t$, UPP predicts future proprioceptive latents over a horizon $\tau_p$:
\begin{equation}
\mathbf{P}^{'}_{t+1:t+\tau_p}
=
f_{\mathrm{upp}}\!\left(
    \mathbf{P}_{t},
    \mathbf{L}'_{t},
    \mathbf{V}'_{t},
    \mathcal{M}^{\mathrm{vl}}_{t}
\right).
\end{equation}

The predicted latents capture short-horizon whole-body evolution and provide future-oriented proprioceptive cues for downstream action generation.

\subsection{Action Expert with Adaptive Fusion}
\label{subsec:action}

As shown in Figure~\ref{fig:upp_model} (b), the action expert generates high-level actions by iteratively denoising action tokens with noise under dual conditioning from visual-language and proprioceptive features. Unlike direct fusion between the two modalities, our action expert uses the evolving action representation itself as the query, and conditions it on both the VLM outputs and the UPP outputs through two parallel cross-attention branches. This design allows action generation to be guided jointly by motion-level vision-semantic context and future-oriented proprioceptive dynamics.

Let $\mathbf{X}_t$ denote the current action hidden states at diffusion step $t$, let $\mathbf{H}^{vl}_t=[\mathbf{L}'_t, \mathbf{V}'_t, \mathcal{M}_t]$ denote the full set of VLM features from the last layer, and let $\mathbf{H}^{p}_t=\mathbf{P}_{t:t+\tau_p}$ denote the proprioceptive features produced by UPP. Action Expert first normalizes the current action states,
\begin{equation}
\widehat{\mathbf{X}}_t = \mathrm{Norm}(\mathbf{X}_t),
\end{equation}
and then applies two cross-attention operations in parallel:
\begin{equation}
\begin{aligned}
\mathbf{H}^{vl}_t
&=
\mathrm{Attn}_{vl}\!\left(
\widehat{\mathbf{X}}_t,\mathbf{H}^{vl}
\right), \\
\mathbf{H}^{p}_t
&=
\mathrm{Attn}_{p}\!\left(
\widehat{\mathbf{X}}_t,\mathbf{H}^{p}
\right).
\end{aligned}
\end{equation}
where the action states serve as queries, while the visual-language and proprioceptive features serve as two conditioning memories.
To combine the two conditioning branches, Action Expert uses a gated fusion mechanism conditioned on both cross-attended features and the current normalized action states:
\begin{equation}
\mathbf{g}_t
=
\sigma\!\left(
W_g
\left[
\mathbf{H}^{vl}_t;
\mathbf{H}^{p}_t;
\widehat{\mathbf{X}}_t
\right]
\right),
\end{equation}
where $\sigma(\cdot)$ is the sigmoid function. The fused conditioning signal is then computed as
\begin{equation}
\mathbf{F}_t
=
\mathbf{H}^{vl}_t
+
\mathbf{g}_t \odot \mathbf{H}^{p}_t,
\end{equation}
and injected into the action states through a residual update:
\begin{equation}
\mathbf{X}'_t = \mathbf{X}_t + \mathbf{F}_t.
\end{equation}
After this dual-conditioning stage, Action Expert further applies self-attention and a feed-forward block to refine the action representation. In this way, the visual-language branch provides semantic and motion-relevant guidance for task execution, while the proprioceptive branch injects future-oriented whole-body dynamics and whole-body coordination cues. The gate adaptively modulates the contribution of the state branch during denoising, enabling the model to generate high-level actions that directly control the arms and hands while remaining consistent with downstream whole-body execution.

\subsection{Cross-Embodiment Training}
\label{sec:training}

We pretrain HEX on trajectory datasets collected from multiple humanoid embodiments, covering diverse kinematics, dynamics, and embodiment-specific state-action spaces. Training jointly optimizes an action-generation objective for the action expert and an auxiliary future-state prediction objective for the Unified Proprioceptive Predictor.

For action generation, we adopt a flow-matching objective. Given a clean future action trajectory $\mathbf{A}_{t:t+\tau_a}$ and Gaussian noise $\mathbf{N}$ of the same shape, we construct a noisy action trajectory
\begin{equation}
\widetilde{\mathbf{A}}_{t:t+\tau_a}
=
(1-\lambda)\mathbf{N} + \lambda \mathbf{A}_{t:t+\tau_a},
\qquad
\lambda \sim \mathcal{U}(0,1),
\end{equation}
and define the corresponding velocity target as
\begin{equation}
\mathbf{Vel}_{t:t+\tau_a}
=
\mathbf{A}_{t:t+\tau_a} - \mathbf{N}.
\end{equation}
Let $\mathbf{Z}^a_t$ denote the final hidden representation produced by the Action Expert, and let $\mathbf{P}_{t+1:t+\tau_p}$ denote the future proprioceptive latents predicted by UPP. The action decoder $D_a(\cdot)$ maps $\mathbf{Z}^a_t$ to velocity predictions, while the state decoder $D_s(\cdot)$ maps $\mathbf{P}_{t+1:t+\tau_p}$ to future proprioceptive states. Let $\mathbf{s}_{t+1:t+\tau_p}$ denote the corresponding ground-truth future proprioceptive trajectory. We then define the training objectives as
\begin{equation}
\begin{aligned}
\mathcal{L}_a
&=
\left\|
D_a(\mathbf{Z}^a_t) - \mathbf{Vel}_{t:t+\tau_a}
\right\|_2^2, \\
\mathcal{L}_s
&=
\left\|
D_s(\mathbf{P}^{'}_{t+1:t+\tau_p})
-
\mathbf{s}_{t+1:t+\tau_p}
\right\|_2^2, \\
\mathcal{L}
&=
\mathcal{L}_a + \alpha\mathcal{L}_s,
\end{aligned}
\end{equation}
where $\mathcal{L}_a$ supervises high-level action denoising under dual conditioning from visual-language and proprioceptive features, while $\mathcal{L}_s$ encourages UPP to model short-horizon proprioceptive evolution in the shared latent space. Because both objectives are defined over the shared body-part-aligned representation, the same training formulation naturally extends across heterogeneous humanoid embodiments. In practice, we optionally adopt a staged schedule for optimization stability: we first warm up UPP using $\mathcal{L}_s$, and then jointly optimize UPP and the Action Expert under the combined objective.

\section{Experiments}
\label{sec:exp}

We conduct extensive experiments on real-world humanoid whole-body manipulation to evaluate the effectiveness, generalization ability, and practical behavior of HEX. In particular, we aim to answer the following four questions:

\begin{itemize}
  
  \item  \textbf{RQ1:} To what extent does HEX improve performance over strong state-of-the-art baselines on real-world humanoid whole-body manipulation, particularly in seen and long-horizon settings? (Section~\ref{subsec:sota})

  \item  \textbf{RQ2:} How effectively does HEX generalize under unseen scene variations? (Section~\ref{subsec:generalization})

  \item  \textbf{RQ3:} What is the effect of each major component in HEX on the overall performance? (Section~\ref{subsec:ablation})

  \item  \textbf{RQ4:} What insights can be drawn from HEX regarding expert routing behavior, inference efficiency, and failure modes? (Section~\ref{subsec:anayses})

\end{itemize}

\subsection{Experiment Setup}

\begin{wrapfigure}{r}{0.35\columnwidth}
    \centering
    \vspace{-8pt}
    \includegraphics[width=\linewidth]{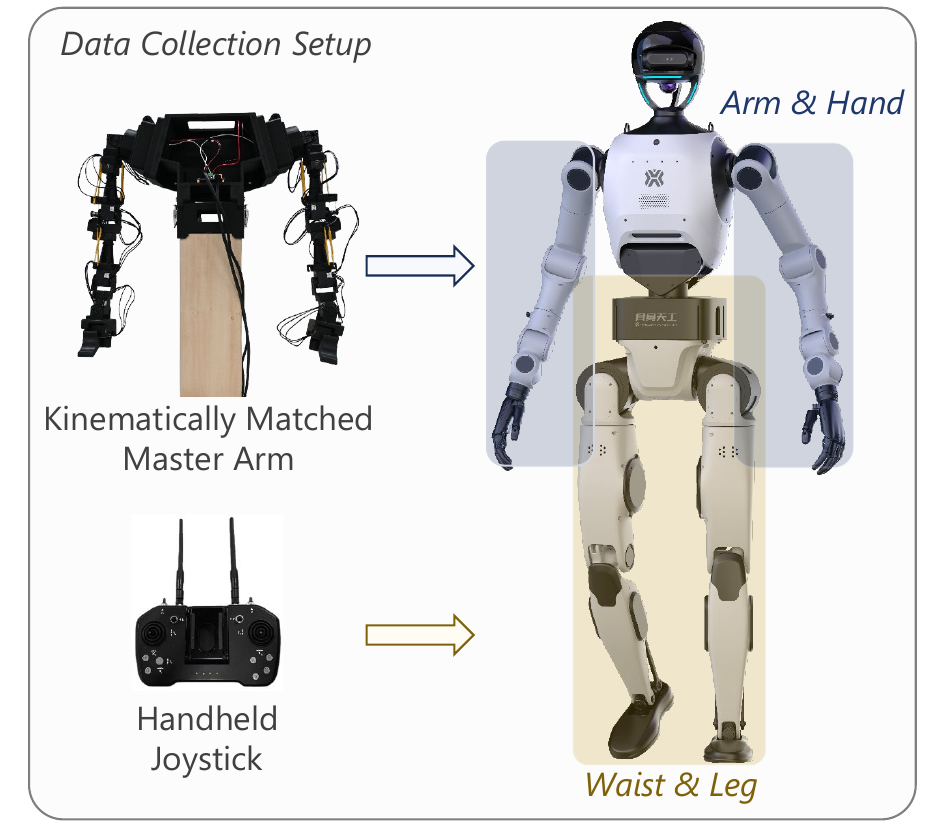}
    \vskip -0.08in
    \caption{Hardware setup for real-robot teleoperation data collection.}
    \label{fig:hardware}
    \vspace{-12pt}
\end{wrapfigure}

\noindent \textbf{Hardware and data collection.}
As shown in Figure~\ref{fig:hardware}, we adopt a modular teleoperation pipeline for data collection. 
Head motion is controlled by computer-issued commands that regulate upward and downward pitch under a fixed data collection protocol. Arm and hand motions are teleoperated through an isomorphic arm--hand interface~\cite{xu2025hacts}, while waist and leg motions are controlled using a handheld joystick. To evaluate cross-embodiment capability, we use two humanoid platforms, TianGong 2.0 and TianGong 3.0. Our data collection procedure follows RoboMIND~\cite{wu2025robomind,hou2025robomind}.

\noindent \textbf{Baselines.}
To ensure a fair comparison of high-level VLA policies, we use the same RL-based low-level controller for balance control across all methods, thereby isolating the contribution of the high-level policy. All models receive the same visual observations, robot proprioceptive information, and language instructions, while preserving their original input encoding and fusion mechanisms. We compare HEX with the following IL and VLA baselines. Unless otherwise specified, all remaining hyperparameters follow the original implementations.

ACT~\cite{zhao2023learning} is a small-scale vision--action model that combines a Transformer encoder--decoder with action chunking. We set the action horizon to 200 and train it for 35k steps on 8 NVIDIA A100 GPUs with a per-GPU batch size of 64.

SwitchVLA~\cite{li2025switchvla} is a medium-scale VLA framework for execution-aware task switching under changing instructions. We set the action horizon to 200 and train it for 100 epochs on 8 NVIDIA A100 GPUs with a per-GPU batch size of 200.

GR00T N1.5~\cite{bjorck2025gr00t} is a large-scale humanoid VLA model trained on both real-world teleoperation and large-scale simulated data. It predicts action sequences from the final-layer VLM features. We set the action horizon to 100 and train it for 50k steps on a single NVIDIA A100 GPU with a batch size of 64.

$\pi_{0.5}$~\cite{intelligence2025pi05} is a large-scale general-purpose robot foundation model in which the VLM and action expert share the same attention backbone. We set the action horizon to 100 and train it for 50k steps on 8 NVIDIA A100 GPUs with a per-GPU batch size of 8.

$\psi_0$~\cite{wei2026psi0} freezes a pretrained vision-language model and trains a separate flow-matching action expert. We set the action horizon to 30 and train it for 50k steps on 8 NVIDIA A100 GPUs with a batch size of 8.

\noindent \textbf{Pretraining Datasets.}
As shown in Figure~\ref{fig:teaser} (a), our training corpus comprises over 12M frames collected from seven humanoid embodiments across four data sources. First, our in-house HEX dataset contains approximately 4M frames from three embodiments: the legged humanoids TianGong 2.0 and TianGong 3.0, and the wheeled humanoid TianYi. Owing to differences in data collection protocols and system versions, the state and action definitions are not fully consistent across embodiments. For example, the state of TianGong 2.0 includes upper- and lower-body proprioceptive signals, while TianGong 3.0 may additionally incorporate IMU measurements and hand tactile signals. Second, the Humanoid Everyday dataset~\cite{zhao2026humanoid} provides approximately 3.4M frames from the legged humanoids Unitree G1 and H1. Its state representation includes both upper- and lower-body information, whereas the action space contains only upper-body actions. Third, AgiBot World Colosseo~\cite{bu2025agibot} contributes 3.8M frames from a wheeled AgiBot humanoid platform. We use its G1-retargeted version~\cite{liu2026trajbooster}, in which the original actions are transformed into a format executable by legged humanoids. Finally, we include 2.3M frames from the Leju legged humanoid subset of RoboCOIN~\cite{wu2025robocoin}. Although these datasets differ substantially in embodiment, state composition, and action parameterization, they can all be leveraged for pretraining within our cross-embodiment architecture.

\noindent \textbf{Implementation Details.}
HEX is built on the vision-language model Qwen3-VL-2B-Instruct~\cite{bai2025qwen3}. The UPP is a 4-layer transformer with hidden size 768, forecasting a 50-step future state horizon. To model embodiment-specific dynamics, we employ a MoE module~\cite{du2025himoe} with 16 routed experts and 2 shared experts, using top-1 softmax routing and an auxiliary load-balancing loss with weight 0.01. The action head is a 16-layer DiT-B with hidden size 1024, which predicts 100-step action chunks conditioned on visual--language features and predicted future states.
During pretraining, HEX is trained for 200k steps with a per-device batch size of 16 and an action chunk size of 100, requiring approximately 1K A100 GPU hours. Optimization is performed using AdamW, with learning rates of $1.0\times10^{-5}$ for the VLM and $2.0\times10^{-5}$ for both the UPP and action modules. We adopt a cosine learning rate schedule with a minimum learning rate of $10^{-6}$, using 5k warmup steps for the main model and 2k warmup steps for the UPP model.
For fine-tuning, each task is trained for 20k steps using AdamW. The learning rate is set to $1.0\times10^{-5}$ for the Qwen-VL interface and $4.0\times10^{-5}$ for both the UPP and action modules. The warmup steps for the UPP model are reduced to 1k. 

During inference, linear interpolation is applied only to the directly predicted arm and hand actions to improve motion smoothness. Lower-body outputs instead serve as commands to the RL-based whole-body controller, which performs closed-loop control and dynamic stabilization at a higher control frequency. We therefore do not additionally interpolate these commands, avoiding unnecessary latency that could reduce responsiveness during balance and locomotion.

\begin{table*}[t]
\centering
\small
\caption{Task success rates (\%) of different methods on TianGong 3.0 and TianGong 2.0 tasks. The icons denote the main body parts involved in control for each task: arm (\includegraphics[width=0.25cm]{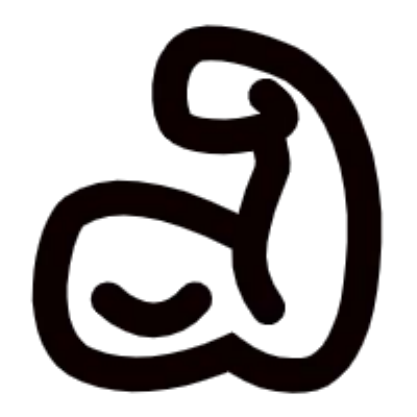}), hand (\includegraphics[width=0.25cm]{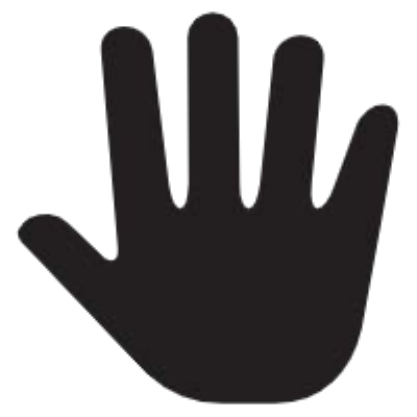}), waist (\includegraphics[width=0.3cm]{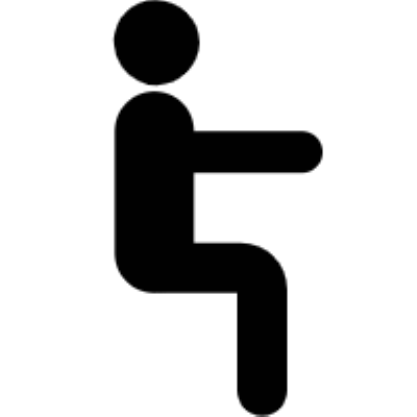}), and leg (\includegraphics[width=0.3cm]{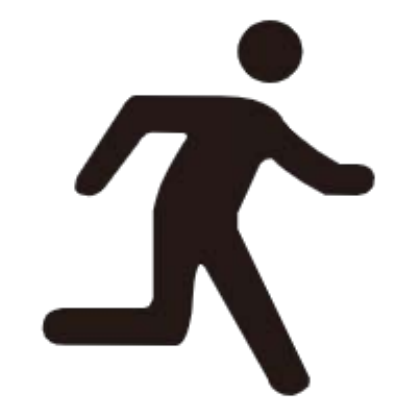}).}
\label{tab:id_exp}
\vskip -0.05in
\resizebox{\linewidth}{!}{
\begin{tabular}{lccccc}
\toprule
\multirow{3}{*}{Method} & \multirow{3}{*}{Para.} & \multirow{3}{*}{Avg (\%)} 
& \multicolumn{1}{c}{TianGong 3.0 (\includegraphics[width=0.25cm]{svgs/arm.pdf} \includegraphics[width=0.25cm]{svgs/hand.pdf} \includegraphics[width=0.3cm]{svgs/waist.pdf})} 
& \multicolumn{1}{c}{TianGong 3.0 (\includegraphics[width=0.25cm]{svgs/arm.pdf} \includegraphics[width=0.25cm]{svgs/hand.pdf} \includegraphics[width=0.3cm]{svgs/waist.pdf})} 
& \multicolumn{1}{c}{TianGong 3.0 (\includegraphics[width=0.25cm]{svgs/arm.pdf} \includegraphics[width=0.3cm]{svgs/waist.pdf} \includegraphics[width=0.3cm]{svgs/leg.pdf})} \\
\cmidrule(lr){4-6}
& & 
& \makecell[c]{Kneel and manipulate \\ the objects}
& Tidy Table
& \makecell[c]{Bring box and \\ pack all objects} \\
\midrule
ACT~\cite{zhao2023learning}        & 80M  & 57.1 & 83.3  & 8.3  & 8.3  \\
SwitchVLA~\cite{li2025switchvla}  & 0.3B & 40.5 & 0.0   & 8.3  & 0.0  \\
GR00T N1.5~\cite{bjorck2025gr00t} & 3B   & 70.2 & 100.0 & 41.7 & 33.3 \\
$\pi_{0.5}$~\cite{intelligence2025pi05} & 3.3B & 71.8 & 100.0 & 35.7 & 25.0 \\
$\psi_0$~\cite{wei2026psi0} & -- & 77.4 & 100.0 & 41.7 & 33.3 \\
HEX & 2.4B & \textbf{79.8} & 100.0 & 41.7 & 41.7 \\
\midrule
\multirow{2}{*}{Method} & \multirow{2}{*}{Para.}
& \multicolumn{1}{c}{TianGong 2.0 (\includegraphics[width=0.25cm]{svgs/arm.pdf} \includegraphics[width=0.3cm]{svgs/waist.pdf})} 
& \multicolumn{1}{c}{TianGong 2.0 (\includegraphics[width=0.25cm]{svgs/arm.pdf} \includegraphics[width=0.25cm]{svgs/hand.pdf} \includegraphics[width=0.3cm]{svgs/waist.pdf})} 
& \multicolumn{1}{c}{TianGong 2.0 (\includegraphics[width=0.25cm]{svgs/arm.pdf} \includegraphics[width=0.3cm]{svgs/waist.pdf} \includegraphics[width=0.3cm]{svgs/leg.pdf})} 
& \multicolumn{1}{c}{TianGong 2.0 (\includegraphics[width=0.25cm]{svgs/arm.pdf} \includegraphics[width=0.3cm]{svgs/waist.pdf} \includegraphics[width=0.3cm]{svgs/leg.pdf})} \\
\cmidrule(lr){3-6}
& 
& Mirror the human's pose
& \makecell[c]{Pour liquor while \\ following the human order}
& Human assistant
& \makecell[c]{Walking while \\ avoiding obstacles} \\
\midrule
ACT~\cite{zhao2023learning}        & 80M  & 83.3  & 83.3  & 66.7  & 66.7  \\
SwitchVLA~\cite{li2025switchvla}  & 0.3B & 100.0 & 41.7  & 58.3  & 75.0  \\
GR00T N1.5~\cite{bjorck2025gr00t} & 3B   & 83.3  & 66.7  & 66.7  & 100.0 \\
$\pi_{0.5}$~\cite{intelligence2025pi05} & 3.3B & 83.3  & 91.7  & 75.0  & 91.7  \\
$\psi_0$~\cite{wei2026psi0} & 2.5B & 100.0 & 91.7 & 75.0 & 100.0 \\
HEX                                & 2.4B & 100.0 & 91.7  & 83.3  & 100.0 \\
\bottomrule
\end{tabular}
}
\end{table*}

\noindent \textbf{Metric.}
For real-world evaluation, task success rate (SR) is used to measure strict end-to-end task completion. However, for long-horizon tasks, binary SR cannot distinguish failures occurring at different intermediate stages. To provide a more fine-grained evaluation, a Stage-wise Progress Score (SPS) is additionally introduced to quantify partial task completion.
Specifically, each long-horizon task is decomposed into $K$ ordered semantic stages, with a predefined weight $w_k$ assigned to stage $k$. Let $I_{i,k}=1$ if evaluation episode $i$ successfully completes stage $k$, and $0$ otherwise. SPS is defined as
\begin{equation}
\mathrm{SPS} =
\frac{100}{N\sum_{k=1}^{K} w_k}
\sum_{i=1}^{N}\sum_{k=1}^{K} w_k I_{i,k},
\end{equation}
where $N$ denotes the number of evaluation episodes. SPS ranges from $0$ to $100$, assigning partial credit to completed intermediate stages, while SR measures strict end-to-end success. The stage definitions and weights are fixed for each task and applied identically to all evaluated methods. 

The stage definitions and weights are fixed for each task and applied identically to all methods. Specifically, \emph{Basket transfer from bed to table}, \emph{Toy transfer to cart basket}, and \emph{Long-horizon box convey} each comprise four stages: grasping, turning, locomotion, and placement, with weights $(3,2,3,2)$, $(3,2,4,3)$, and $(3,2,4,3)$, respectively. \emph{Pouring} comprises five stages: grasping, pouring into three cups, and returning the arm, with weights $(3,3,3,3,2)$.

\subsection{Comparing with SoTA}
\label{subsec:sota}

\subsubsection{Seen Scenarios}
Seen scenarios refer to evaluation settings where the test environments closely match those in the training data. This setting primarily assesses the ability of VLA models to reproduce demonstrated trajectories from observations. All methods are evaluated over 12 trials.

\noindent \textbf{Post-training Datasets.}
We collect seven real-robot tasks in total, including four on Tiangong 2.0 and three on Tiangong 3.0. These tasks cover whole-body control involving the arms and hands, waist, and legs, as well as multiple scenarios requiring timely human-robot interaction, enabling evaluation of both task success rate and response speed across different VLA models. The seven tasks are as follows.

Task 1: Mirror the human's pose. The robot observes a person standing in front of it and imitates the posed gestures, including “V,” “L,” and “A,” in real time. We collect 108 trajectories for training.

Task 2: Pour liquor while following human order. A liquor bottle and three cups are placed on the table in front of the robot. The robot pours liquid into the cup indicated by the human's pointing. We collect 100 trajectories for training.

Task 3: Human assistant. The robot carries a box and follows a human collaborator to assist with organizing objects across two tables. We collect 98 trajectories for training.

Task 4: Walking while avoiding obstacles. As the robot walks forward, it must stop promptly when a person or cart passes through its path, and then resume walking once the path is clear. We collect 100 trajectories for training.

Task 5: Kneel and manipulate the objects. The robot kneels down to pick up blocks and place them into a box. We collect 300 trajectories for training.

Task 6: Tidy table. The robot clears the tabletop by sorting scattered objects into a box and disposing of paper scraps into a trash bin. We collect 100 trajectories for training.

Task 7: Bring box and pack all objects. The robot first retrieves a box and then packs all target objects into it. We collect 100 trajectories for training.

\noindent \textbf{Results.}
As shown in Table~\ref{tab:id_exp}, in in-distribution settings, despite their much smaller parameter scales, ACT and SwitchVLA remain competitive with several-billion-parameter models, indicating that relatively compact policies can already achieve strong performance on seen tasks. In particular, ACT exhibits strong trajectory-following behavior and produces remarkably smooth hand motions on several reactive tasks with almost no observable latency. Among the large-scale models, $\pi_{0.5}$ slightly outperforms GR00T N1.5 in average success rate, while $\psi_0$ further improves overall performance and HEX achieves the best average success rate of 79.8\%. Qualitatively, large-scale models may exhibit mild lag or stuttering in highly reactive tasks, whereas smaller policies such as ACT often produce smoother motions. Overall, HEX provides the strongest balance between task success and motion quality among the evaluated large-scale models, which may benefit from its explicit future-state conditioning and predictive proprioceptive modeling.

\begin{table*}[t]
\centering
\small
\caption{Task success rate (SR, \%) and stage-wise progress score (SPS) of different methods on three long-horizon whole-body manipulation tasks.}
\label{tab:long_horizon}
\vskip -0.05in
\resizebox{0.87\linewidth}{!}{
\begin{tabular}{lccccccc}
\toprule
\multirow{3}{*}{Method}
& \multirow{3}{*}{Para.}
& \multicolumn{2}{c}{TianGong 2.0 (\includegraphics[width=0.25cm]{svgs/arm.pdf} \includegraphics[width=0.25cm]{svgs/hand.pdf} \includegraphics[width=0.3cm]{svgs/waist.pdf} \includegraphics[width=0.3cm]{svgs/leg.pdf})}
& \multicolumn{4}{c}{TianGong 3.0 (\includegraphics[width=0.25cm]{svgs/arm.pdf} \includegraphics[width=0.25cm]{svgs/hand.pdf} \includegraphics[width=0.3cm]{svgs/waist.pdf} \includegraphics[width=0.3cm]{svgs/leg.pdf})} \\
\cmidrule(lr){3-4}
\cmidrule(lr){5-8}

& &
\multicolumn{2}{c}{Long-horizon box convey}
& \multicolumn{2}{c}{Basket transfer from bed to table}
& \multicolumn{2}{c}{Toy transfer to cart basket} \\
\cmidrule(lr){3-4}
\cmidrule(lr){5-6}
\cmidrule(lr){7-8}

& &
SR (\%) & SPS
& SR (\%) & SPS
& SR (\%) & SPS \\
\midrule

ACT~\cite{zhao2023learning}
& 80M
& 26.7 & 55.6
& 6.7 & 18.9
& 0.0 & 18.0 \\

SwitchVLA~\cite{li2025switchvla}
& 0.3B
& 13.3 & 42.8
& 0.0 & 5.0
& 0.0 & 16.0 \\

GR00T N1.5~\cite{bjorck2025gr00t}
& 3B
& 20.0 & 47.8
& 60.0 & 70.0
& 33.3 & 56.7 \\

$\pi_{0.5}$~\cite{intelligence2025pi05}
& 3.3B
& 40.0 & 76.1
& 46.7 & 62.2
& 60.0 & 77.3 \\

$\psi_0$~\cite{wei2026psi0}
& 2.5B
& 46.7 & 75.6
& 46.7 & 67.2
& 60.0 & 74.0 \\

HEX
& 2.4B
& \textbf{53.3} & \textbf{79.4}
& \textbf{73.3} & \textbf{83.9}
& \textbf{86.7} & \textbf{92.0} \\

\bottomrule
\end{tabular}
}
\end{table*}

\subsubsection{Long-Horizon Scenarios}
Long-horizon tasks are composed of multiple subtasks, where different stages require different body parts, such as the waist and hands in some stages and the legs in others. This heterogeneous composition increases task complexity and gives rise to more pronounced cascading errors across stages.

\noindent \textbf{Post-training Datasets.}
We collect a long-horizon box conveyance task consisting of four stages: squatting down to grasp the box, turning toward the table, moving to the table and stopping in front of it, and squatting down again to place the box. In total, we collect 56 trajectories for training.
We further introduce two household long-horizon tasks, \emph{Basket transfer from bed to table} and \emph{Toy transfer to cart basket}, each with 100 post-training trajectories. Both tasks require coordinated manipulation and locomotion over multiple stages, including object grasping, body turning, object transport, and final placement.
Each task is evaluated over 15 trials.

\noindent \textbf{Results.}
As shown in Table~\ref{tab:long_horizon}, HEX consistently achieves the best SR and SPS across all three long-horizon tasks. On the original box convey task, HEX reaches 53.3\% SR and 79.4 SPS. The advantage becomes more pronounced on the two additional tasks, achieving 73.3\% SR and 83.9 SPS on basket transfer, and 86.7\% SR and 92.0 SPS on toy transfer. Compared with the strongest baseline on each task, HEX improves SR by 6.6, 13.3, and 26.7 percentage points, respectively. These results demonstrate that HEX more reliably maintains task progress across multiple manipulation and locomotion stages, reducing performance degradation over long-horizon whole-body execution.

\begin{figure*}[t]
\centering
\includegraphics[width=0.95\textwidth]{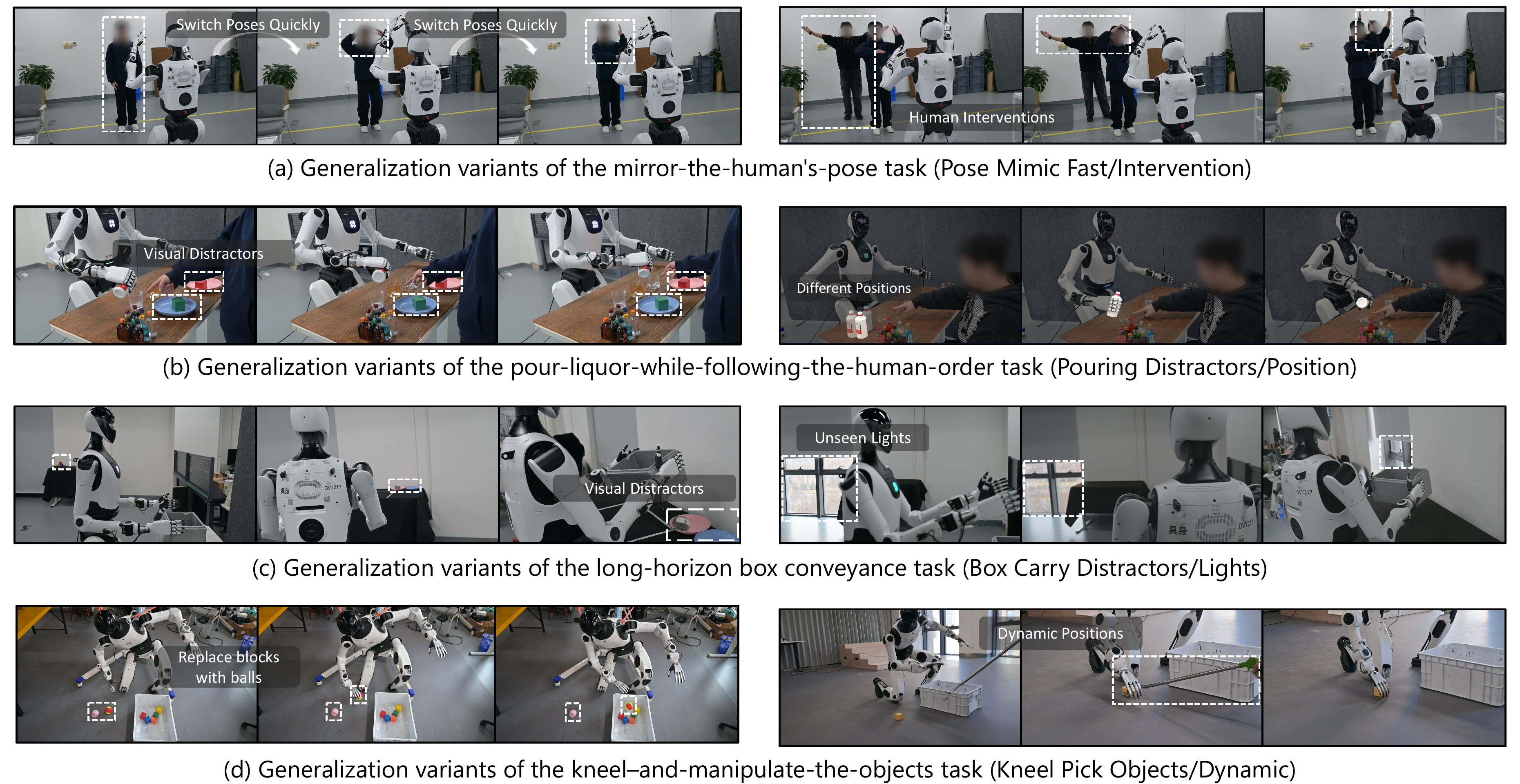}
\vskip -0.1 in
\caption{Generalization tasks. Two distribution-shift variants for each of four seen tasks: \textit{Pose Mimic}, \textit{Pouring}, \textit{Box Carry}, and \textit{Kneel Pick}.
}
\label{fig:generalization_tasks}
\end{figure*}

\subsection{Generalization Study}
\label{subsec:generalization}

\noindent \textbf{Evalation Tasks.}
As shown in Figure~\ref{fig:generalization_tasks}, we evaluate generalization on four tasks from the seen-scenario setting, including three standard tasks (Tasks 1, 2, and 5) and one long-horizon task. For \textit{Pose Mimic}, we consider Pose Mimic Fast, which increases the speed of human pose switching, and Pose Mimic Intervention, where an additional person in the background continuously performs distracting poses. A total of 18 trials are conducted, including 5 trials each for the V-, L-, and A-shaped poses, and 3 trials for the return-hand pose.
For \textit{Pouring}, we evaluate Pouring Distractors by adding irrelevant objects, and Pouring Position by changing the bottle location. In the distractor setting, three cups are tested with 5 trials each. In the position-variation setting, the bottle is placed at 9 different locations, with one trial for each cup at each location.
For \textit{Kneel Pick}, we evaluate Kneel Pick Dynamic, where object positions are changed during execution, and Kneel Pick Objects, where the original blocks are replaced with unseen balls. Each variant is evaluated with 15 grasp-and-place trials.
For \textit{Box Carry}, we evaluate Box Carry Distractors by introducing additional surrounding objects, and Box Carry Lights by changing the lighting condition. Each variant is evaluated over 15 trials.

\begin{wrapfigure}{r}{0.5\columnwidth}
    \centering
    \vspace{-8pt}
    \includegraphics[width=\linewidth]{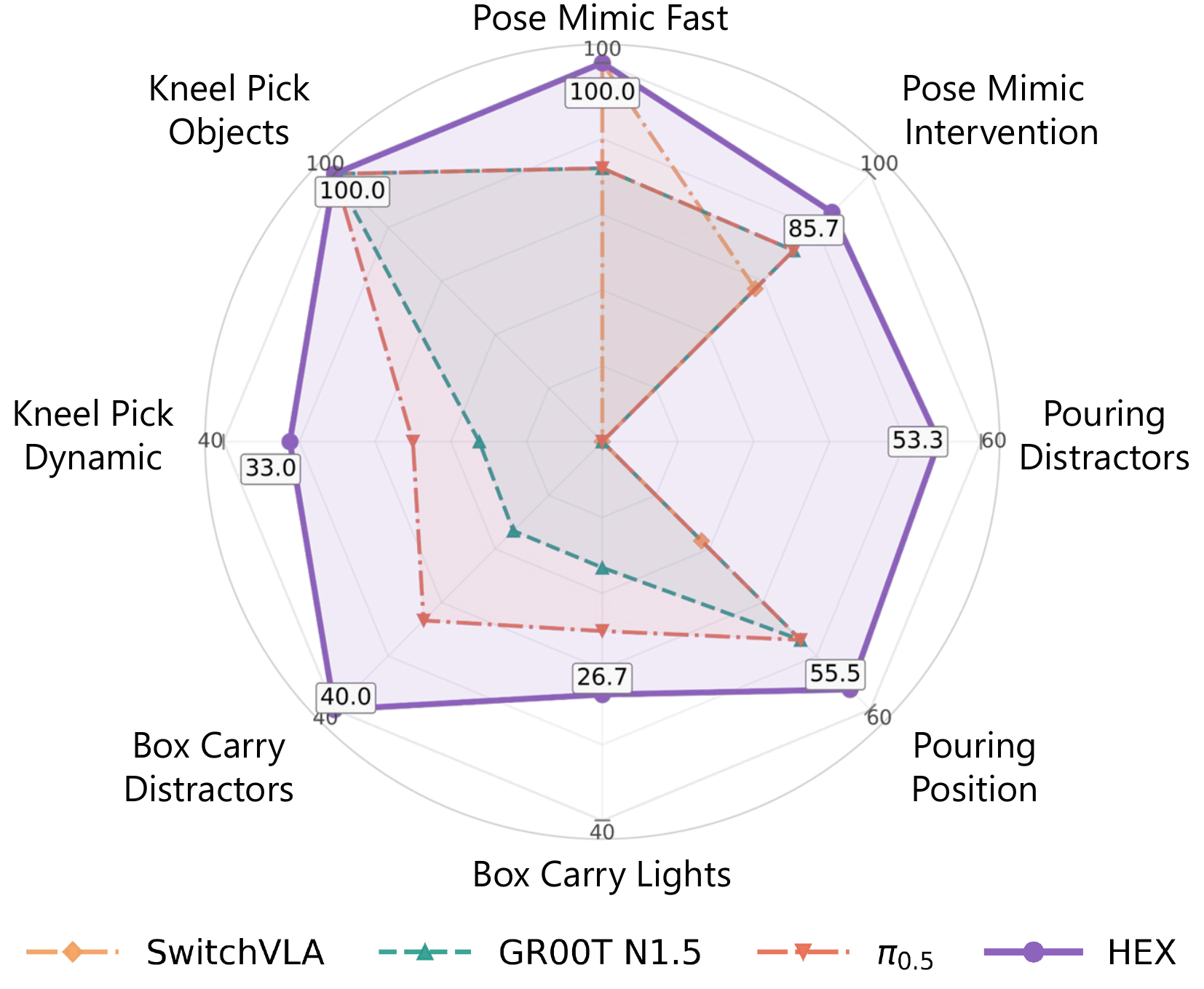}
    \vskip -0.08in
    \caption{
    Generalization results across unseen task variants.
    }
    \label{fig: generalization}
    \vspace{-8pt}
\end{wrapfigure}
\noindent \textbf{Results.}
Figure~\ref{fig: generalization} summarizes the results on eight generalization variants across four seen tasks. 
HEX achieves the best overall average success rate of 61.8\%, substantially outperforming $\pi_{0.5}$ (44.3\%), GR00T N1.5 (41.0\%), and SwitchVLA (22.4\%). Overall, HEX performs best on nearly all variants.
In \textit{Pose Mimic}, HEX matches the best result under fast switching (100\%) and achieves the highest success rate under human intervention (85.7\%). Notably, under human intervention, all methods except HEX fail to maintain the ``L'' hand gesture after being distracted by a background person. In addition, GR00T N1.5 and $\pi_{0.5}$ often produce inaccurate \texttt{L}-shaped poses. In contrast, HEX remains substantially more robust to such interference.
In \textit{Pouring}, HEX shows the clearest advantage, improving from 0\% for all baselines to 53.3\% under visual distractors, and reaching 55.5\% under bottle-position changes. After distractor objects are introduced, all baselines tend to start pouring before receiving a human instruction and then remain fixed at one location. We conjecture that these models mistakenly treat the red plate as the human hand, whereas HEX does not exhibit this failure mode.
In \textit{Box Carry}, HEX also achieves the best results under both unseen lighting (26.7\%) and unseen surrounding objects (40.0\%). For \textit{Kneel Pick}, HEX attains 33.0\% under dynamic position changes and 100\% under unseen objects.
These results indicate that HEX generalizes more robustly under diverse distribution shifts, including faster human motion, human interference, visual distractors, object-position changes, lighting variation, and dynamic scene changes.

\subsection{Ablation Study}
\label{subsec:ablation}

\subsubsection{Ablation on Pretraining}
We evaluate the impact of pretraining from two perspectives: in-distribution optimization efficiency and out-of-distribution generalization.

\begin{wrapfigure}{r}{0.55\columnwidth}
    \centering
    \vspace{-8pt}
    \includegraphics[width=\linewidth]{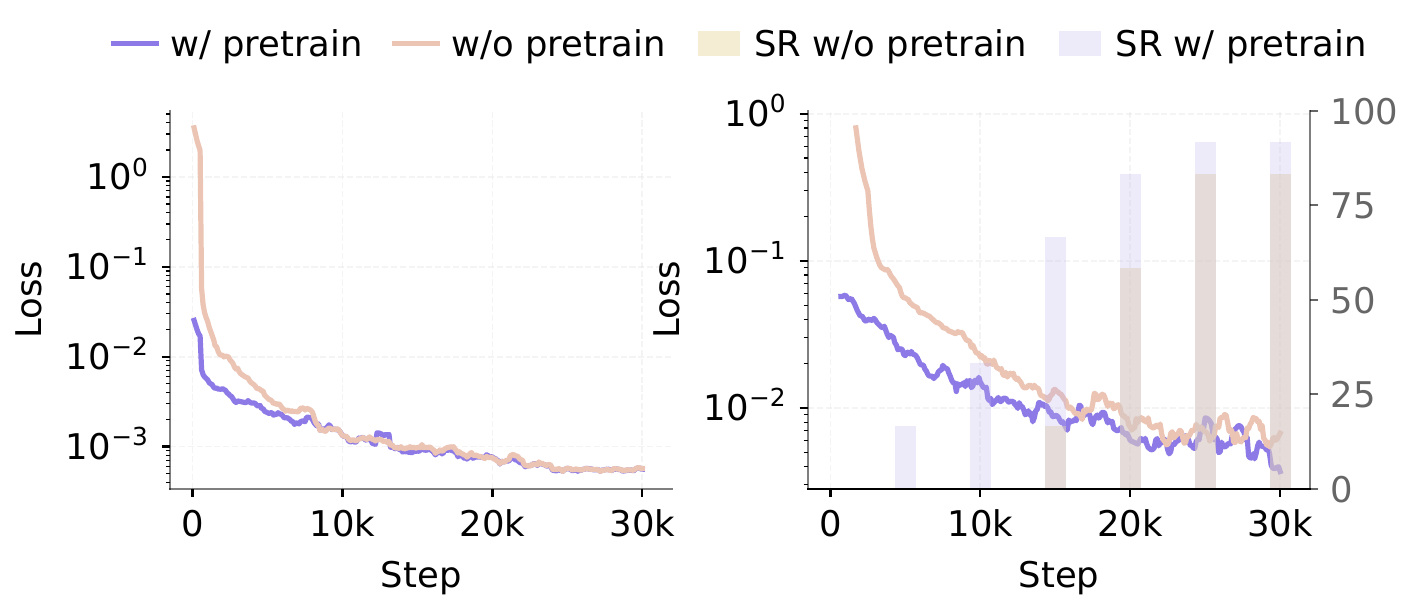}
    \vskip -0.08in
    \caption{Effect of pretraining. Left: state loss. Right: action loss together with success-rate comparisons at different training stages. 
    }
    \label{fig:ablation_pretrain}
    \vspace{-8pt}
\end{wrapfigure}
\noindent \textbf{In-distribution Performance.}
Figure~\ref{fig:ablation_pretrain}(a) shows that pretraining mainly improves optimization efficiency rather than final converged performance in the single-task setting. Specifically, the pretrained model exhibits lower state and action losses during early training, indicating better initialization and faster convergence. As training proceeds, the state loss gap becomes marginal after approximately 10k steps, while the pretrained model generally maintains a lower action loss. This advantage is also reflected in early-stage task success: at 5k/10k/15k/20k steps, the pretrained model achieves 2/12, 4/12, 8/12, and 10/12 successes, compared with 0/12, 0/12, 2/12, and 7/12 without pretraining. However, both models eventually achieve comparable success rates (11/12 vs.\ 10/12). These results suggest that the primary benefit of pretraining under in-distribution conditions lies in faster optimization and improved training efficiency, rather than substantially higher final performance.

\begin{wrapfigure}{r}{0.55\columnwidth}
    \centering
    \vspace{-12pt}
    \includegraphics[width=\linewidth]{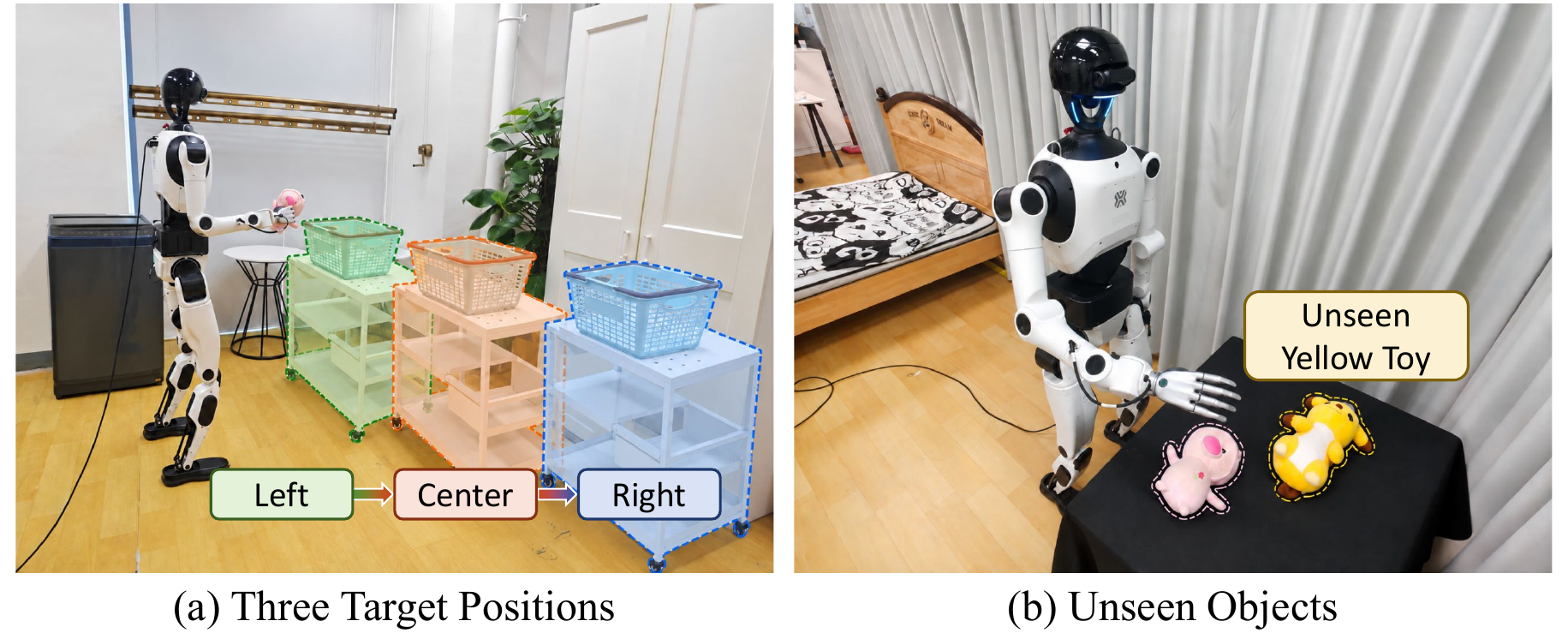}
    \vskip -0.08in
    \caption{Real-world generalization settings with and without pretraining. (a) Spatial generalization to unseen cart positions. (b) Object generalization to unseen plush toys.}
    \label{fig:pretrain_ood}
    \vspace{-8pt}
\end{wrapfigure}
\noindent \textbf{Out-of-distribution Generalization.}
To further examine whether pretraining improves generalization beyond faster downstream optimization, we evaluate HEX on the long-horizon \emph{Toy transfer to cart basket} task under two distribution shifts: unseen target cart positions and unseen manipulated objects, as illustrated in Figure~\ref{fig:pretrain_ood}. Both models are evaluated under the same downstream training and testing settings, differing only in whether pretraining is applied. As shown in Table~\ref{tab:pretrain_ood}, pretraining substantially improves performance under both shifts. For unseen target positions, SR increases from 40.0\% to 73.3\%, while SPS improves from 61.7 to 83.9. Without pretraining, the policy often completes initial manipulation but struggles with target-directed locomotion and accurate placement, particularly for distant targets. In contrast, the pretrained model exhibits more consistent navigation and placement across target positions. For unseen objects, the improvement is more pronounced, with SR increasing from 0.0\% to 66.7\% and SPS from 10.6 to 77.2. The non-pretrained model frequently fails to establish stable grasps or execute locomotion, whereas the pretrained model achieves reliable whole-body task execution. These results show that pretraining not only accelerates downstream optimization but also improves robustness to unseen objects and spatial configurations.

\begin{table}[htbp]
\centering
\small
\caption{Effect of pretraining on OOD generalization for the \emph{Toy transfer to cart basket} task. We evaluate two distribution shifts: unseen target positions and an unseen object.}
\label{tab:pretrain_ood}
\vskip -0.05in
\begin{tabular}{l|lcc}
\toprule
OOD Setting & Pretraining & SR (\%) $\uparrow$ & SPS $\uparrow$ \\
\midrule
\multirow{2}{*}{Three target positions}
& w/o Pretrain & 40.0 & 61.7 \\
& w/ Pretrain  & \textbf{73.3} & \textbf{83.9} \\
\midrule
\multirow{2}{*}{Unseen object}
& w/o Pretrain & 0.0 & 10.6 \\
& w/ Pretrain  & \textbf{66.7} & \textbf{77.2} \\
\bottomrule
\end{tabular}
\end{table}

\subsubsection{Data Scaling}
We investigate the effects of data scaling at both the pretraining and fine-tuning stages, examining how pretraining data composition and downstream supervision influence task performance.

\noindent \textbf{Ablation on Pretraining Data Composition.}
We investigate how different pretraining data sources and their combinations affect downstream performance on the long-horizon \emph{Basket transfer from bed to table} task under the same fine-tuning protocol. As shown in Table~\ref{tab:pretrain_data}, HEX and HE-G1 achieve the strongest single-source transfer, both reaching 80.0\% SR, with SPS values of 88.7 and 86.7, respectively. Notably, HE-G1 transfers effectively despite substantial embodiment differences from Tiangong, demonstrating the potential of cross-embodiment knowledge transfer. In contrast, Agibot, HE-H1, and RoboCOIN-Leju yield weaker performance. Uniformly mixing multiple sources does not always help: adding Agibot to HEX reduces SR from 80.0\% to 46.7\%, whereas combining HEX with HE-G1 preserves strong performance. These results suggest that data quality and transferability may matter more than embodiment similarity alone. Accordingly, the final all-source configuration adopts quality-aware sampling, prioritizing stronger sources while retaining weaker ones for diversity. This strategy achieves the best overall performance (86.7\% SR and 90.0 SPS), highlighting the importance of balancing data quality and diversity in cross-embodiment pretraining.

\begin{table}[htbp]
\centering
\small
\caption{Ablation on pretraining data composition for the \emph{Basket transfer from bed to table} task. All models use the same downstream fine-tuning and evaluation setting.}
\label{tab:pretrain_data}
\vskip -0.05in
\begin{tabular}{llcc}
\toprule
Setting & Pretraining Data & SR (\%) $\uparrow$ & SPS $\uparrow$ \\
\midrule
\multirow{5}{*}{Single-source}
& Agibot & 0.0 & 46.7 \\
& HE-G1 & \textbf{80.0} & 86.7 \\
& HE-H1 & 0.0 & 26.0 \\
& RoboCOIN-Leju & 0.0 & 6.7 \\
& HEX data & \textbf{80.0} & \textbf{88.7} \\
\midrule
\multirow{4}{*}{Multi-source}
& HEX + Agibot & 46.7 & 58.7 \\
& HEX + HE-G1 & 80.0 & 88.7 \\
& HEX + HE-H1 & 0.0 & 33.3 \\
& All sources & \textbf{86.7} & \textbf{90.0} \\
\bottomrule
\end{tabular}
\end{table}

\begin{figure*}[t]
\centering
\includegraphics[width=\textwidth]{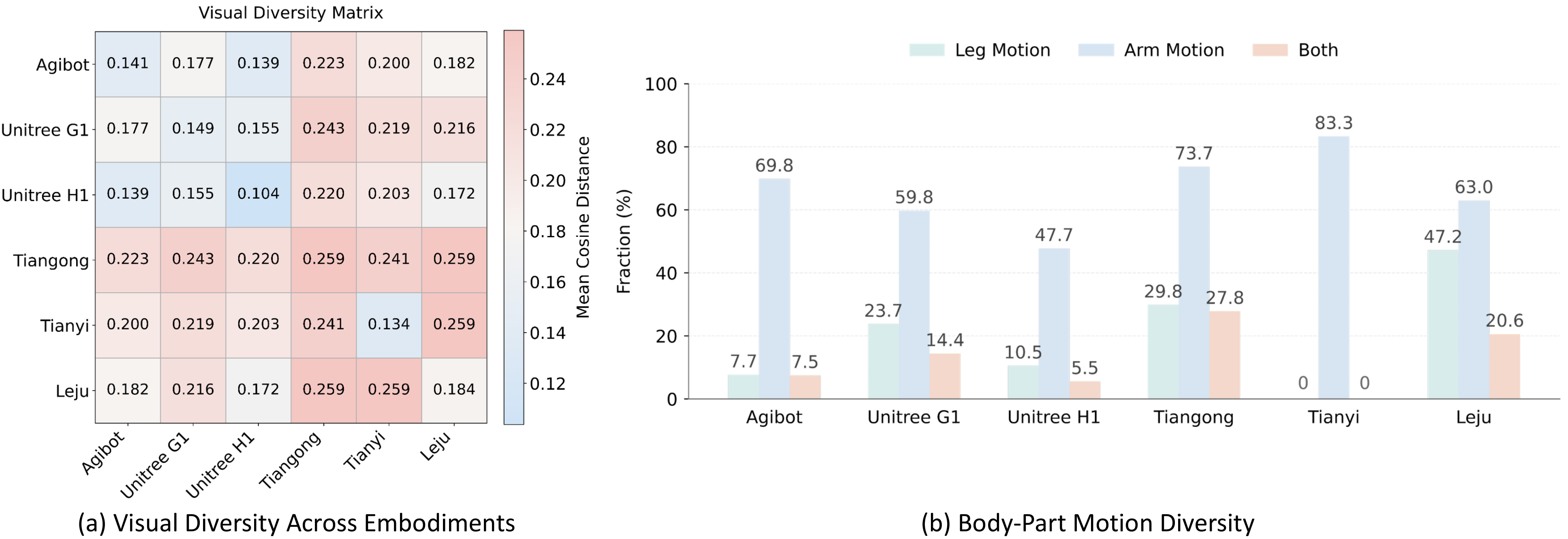}
\vskip -0.1in
\caption{Analysis of visual and motion diversity across pretraining datasets. (a) Mean pairwise cosine distances between visual features, where diagonal and off-diagonal entries represent intra-source diversity and inter-source differences, respectively. (b) Proportions of samples exhibiting leg motion, arm motion, and simultaneous arm-leg motion across different datasets.}
\label{fig:data_diversity}
\end{figure*}

To better understand these performance differences, we further examine three characteristics of the pretraining datasets: task diversity, visual diversity, and whole-body motion coverage. First, Agibot, Unitree G1, Unitree H1, Leju, and Tiangong contain 175, 247, 247, 5, and 34 task categories, respectively. However, task count alone does not determine transferability, as Unitree G1 and H1 exhibit different downstream performance despite having identical task counts. Second, we sample 2K frames from each dataset and measure visual diversity using the mean pairwise cosine distance between extracted features. Figure~\ref{fig:data_diversity}(a) shows that Tiangong exhibits relatively high intra-source and inter-source visual distances, whereas Agibot and Unitree H1 show lower visual variation. Third, Figure~\ref{fig:data_diversity}(b) shows that Tiangong, Unitree G1, and Leju contain higher proportions of coordinated arm-leg motion, while Agibot and Unitree H1 exhibit more limited coverage. This may partly explain their weaker transfer to tasks requiring coordinated locomotion and manipulation. Notably, Leju exhibits relatively high motion coverage but limited task diversity, suggesting that no single data characteristic fully explains downstream transferability. Together, these findings highlight that effective cross-embodiment pretraining requires diverse and task-relevant visual and motion experiences, rather than merely increasing the number of tasks or robot embodiments.

\begin{figure*}[t]
\centering
\includegraphics[width=0.8\textwidth]{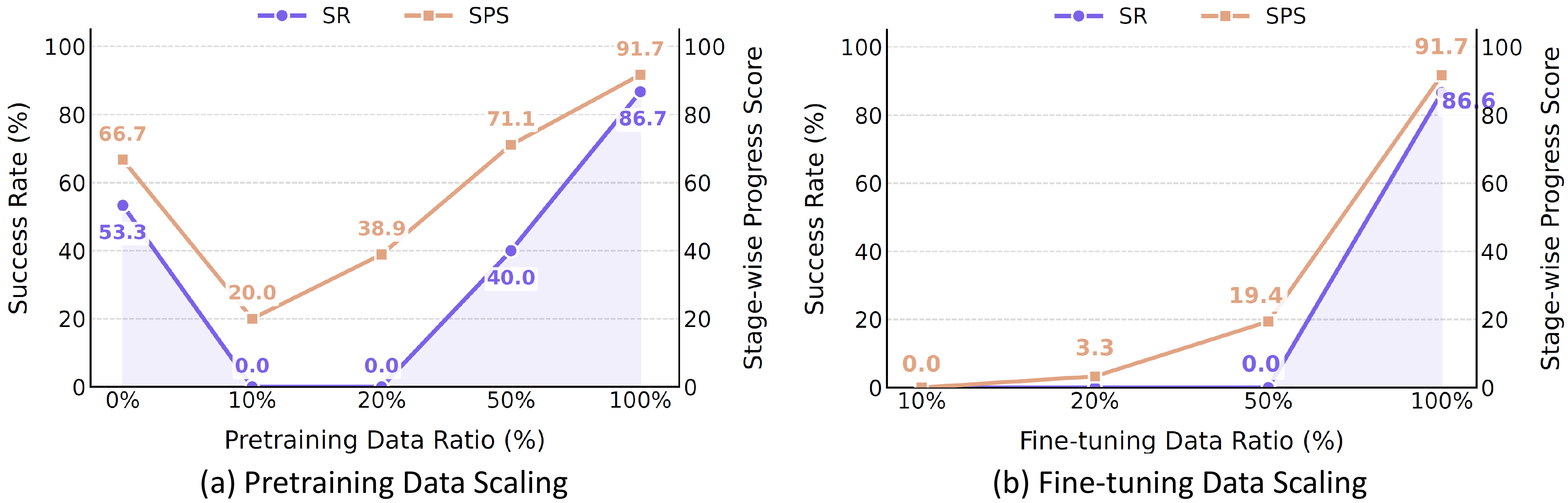}
\vskip -0.1 in
\caption{Effect of pretraining and fine-tuning data ratio on the long-horizon \emph{Basket transfer from bed to table} task.}
\label{fig:data_scale}
\end{figure*}

\noindent \textbf{Ablation on Pretraining Data Scaling.}
To investigate the impact of pretraining data scale, we evaluate HEX using different fractions of the pretraining dataset while keeping the downstream fine-tuning data and protocol unchanged. As shown in Figure~\ref{fig:data_scale}(a), pretraining exhibits a non-monotonic effect on downstream performance. Surprisingly, using only 10-20\% of the pretraining data reduces SR from 53.3\% (training from scratch) to 0.0\%, indicating that insufficient pretraining can introduce negative transfer rather than improve performance. One possible explanation is that limited pretraining data fails to capture sufficiently diverse embodiment-specific dynamics, resulting in biased representations that hinder downstream adaptation. As the pretraining scale increases to 50\%, SPS improves to 71.1, although SR remains below the scratch baseline at 40.0\%. With the full pretraining dataset, HEX achieves 86.7\% SR and 91.7 SPS, substantially outperforming training from scratch. These findings suggest that effective cross-embodiment pretraining requires sufficient data coverage to learn transferable motion representations, as limited heterogeneous data may introduce harmful inductive biases rather than beneficial prior knowledge.

\noindent \textbf{Ablation on Fine-tuning Data Scaling.}
To further investigate the data efficiency of downstream adaptation, we evaluate HEX on the long-horizon \emph{Basket transfer from bed to table} task using different fractions of task-specific training data. As shown in Figure~\ref{fig:data_scale}(b), increasing the fine-tuning data ratio progressively improves task performance. Although no complete task success is achieved with 10-50\% of the training data, the stage-wise progress score (SPS) increases from 0.0 at 10\% to 3.3 at 20\% and 19.4 at 50\%, indicating improved partial task completion. With the full training dataset, HEX achieves a success rate (SR) of 86.6\% and an SPS of 91.7. These results characterize the adaptation behavior of the cross-embodiment pretrained model under varying amounts of downstream supervision, while also highlighting the importance of sufficient task-specific data for successful long-horizon whole-body manipulation.

\subsubsection{Ablation on Model Components}
\begin{figure}[t]
\centering
\includegraphics[width=0.7\textwidth]{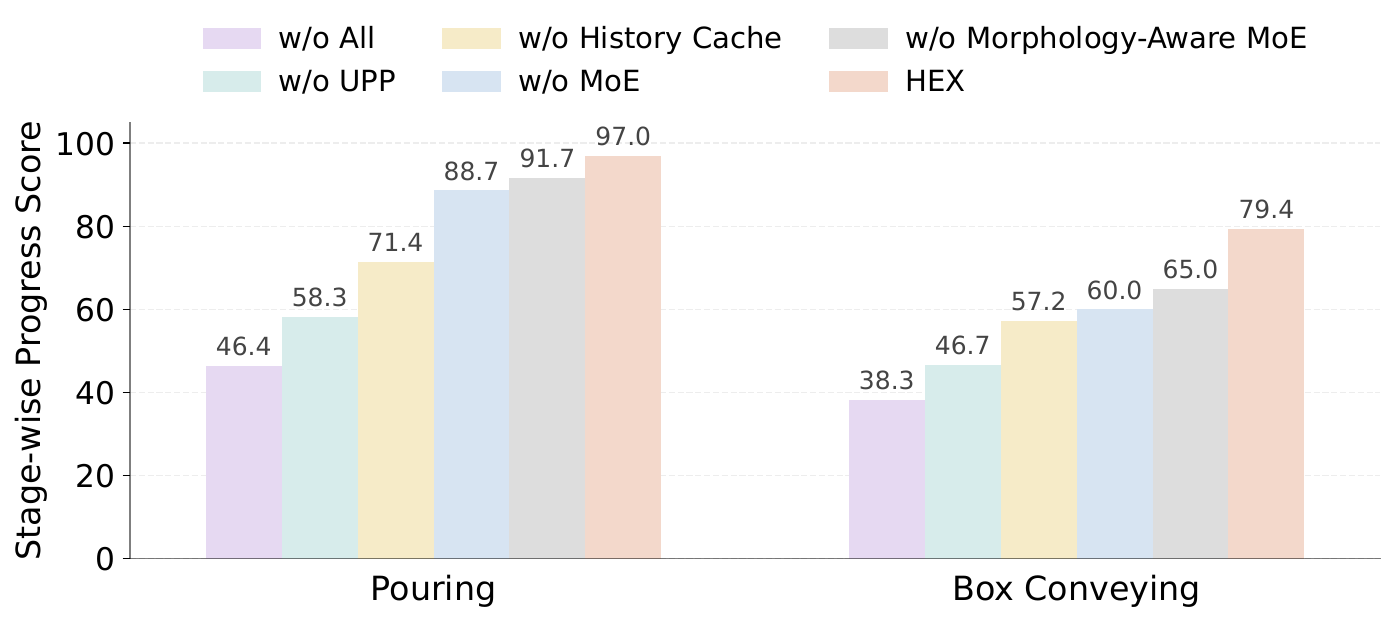}
\vskip -0.12in
\caption{Effect of key model components. Performance improves consistently as the history cache, UPP, and MoE design are progressively introduced, and the full HEX achieves the best SPS on both \textit{Pouring} and \textit{Box Conveying}.}
\label{fig:ablation_component}
\end{figure}

Figure~\ref{fig:ablation_component} evaluates the contributions of the VLM history cache, UPP, and morphology-aware MoE design. Performance consistently improves as these components are progressively introduced, highlighting their complementary roles in complex whole-body manipulation. In particular, UPP contributes substantially to overall performance, underscoring the importance of accurate proprioceptive dynamics modeling for coordinated control. On \emph{Pour} and \emph{Box Conveying}, the morphology-agnostic MoE achieves SPS values of 91.7 and 65.0, outperforming the ``w/o MoE'' variant (88.7 and 60.0). Incorporating morphology-aware routing further improves SPS to 97.0 and 79.4, respectively, under the same model capacity and expert architecture. These results suggest that effective whole-body control benefits not only from specialized expert representations but also from explicitly accounting for embodiment-specific structural differences, enabling more effective modeling of heterogeneous whole-body motion dynamics.

\begin{table*}[htbp]

\centering
\small
\caption{Parameter sensitivity analysis of the visual history window, number of experts in UPP, and loss weight $\alpha$.}
\label{tab:ablation_hyperparams}
\vskip -0.05in

\begin{minipage}[t]{0.31\textwidth}
\centering
\textbf{(a) Visual History Window}
\vskip 0.05in
\begin{tabular}{c|cc}
\toprule
$H$ & SR (\%) $\uparrow$ & SPS $\uparrow$ \\
\midrule
0 & 0.0 & 43.3 \\
2 & \textbf{80.0} & \textbf{85.3} \\
4 & 46.6 & 52.0 \\
8 & 0.0 & 28.7 \\
\bottomrule
\end{tabular}
\end{minipage}
\hfill
\begin{minipage}[t]{0.31\textwidth}
\centering
\textbf{(b) Number of Experts in UPP}
\vskip 0.05in
\begin{tabular}{c|cc}
\toprule
\# Experts & SR (\%) $\uparrow$ & SPS $\uparrow$ \\
\midrule
1  & 13.3 & 40.0 \\
8  & 33.3 & 58.7 \\
16 & \textbf{53.3} & 68.7 \\
32 & \textbf{53.3} & \textbf{72.0} \\
\bottomrule
\end{tabular}
\end{minipage}
\hfill
\begin{minipage}[t]{0.31\textwidth}
\centering
\textbf{(c) Loss Weight $\alpha$}
\vskip 0.05in
\begin{tabular}{c|cc}
\toprule
$\alpha$ & SR (\%) $\uparrow$ & SPS $\uparrow$ \\
\midrule
0.0 & 0.0 & 30.7 \\
0.5 & 73.3 & \textbf{89.3} \\
1.0 & \textbf{80.0} & 85.3 \\
2.0 & 0.0 & 18.7 \\
\bottomrule
\end{tabular}
\end{minipage}

\end{table*}

\subsection{Other Anayses}
\label{subsec:anayses}

\subsubsection{Parameter Sensitivity Analysis}
Table~\ref{tab:ablation_hyperparams} provides a sensitivity analysis of HEX for three key hyperparameters: the visual history window $H$, the number of experts in UPP, and the loss weight $\alpha$ used to balance action and state prediction objectives.

\noindent \textbf{Visual history window.}
Using only the current observation ($H=0$) leads to a substantial performance drop, indicating that temporal visual context is important for long-horizon whole-body execution. A short history window of $H=2$ achieves the best performance, with 80.0\% SR and 85.3 SPS. Increasing the window to $H=4$ or $H=8$ does not yield further improvements. In particular, with $H=8$, the policy tends to over-rotate during the turning stage, which subsequently affects locomotion. This suggests that a compact visual history provides sufficient temporal cues while avoiding excessive reliance on outdated observations. We therefore use $H=2$ by default.

\noindent \textbf{Number of experts.}
Increasing the number of UPP experts from 1 to 16 consistently improves both SR and SPS, demonstrating the benefit of additional expert capacity for modeling heterogeneous whole-body dynamics. Increasing the number further from 16 to 32 yields the same SR (53.3\%) and only a modest SPS improvement from 68.7 to 72.0, while introducing additional computational and memory overhead. We therefore adopt 16 experts as a practical trade-off between performance and efficiency.

\begin{figure*}[!t]
\centering
\includegraphics[width=0.9\textwidth]{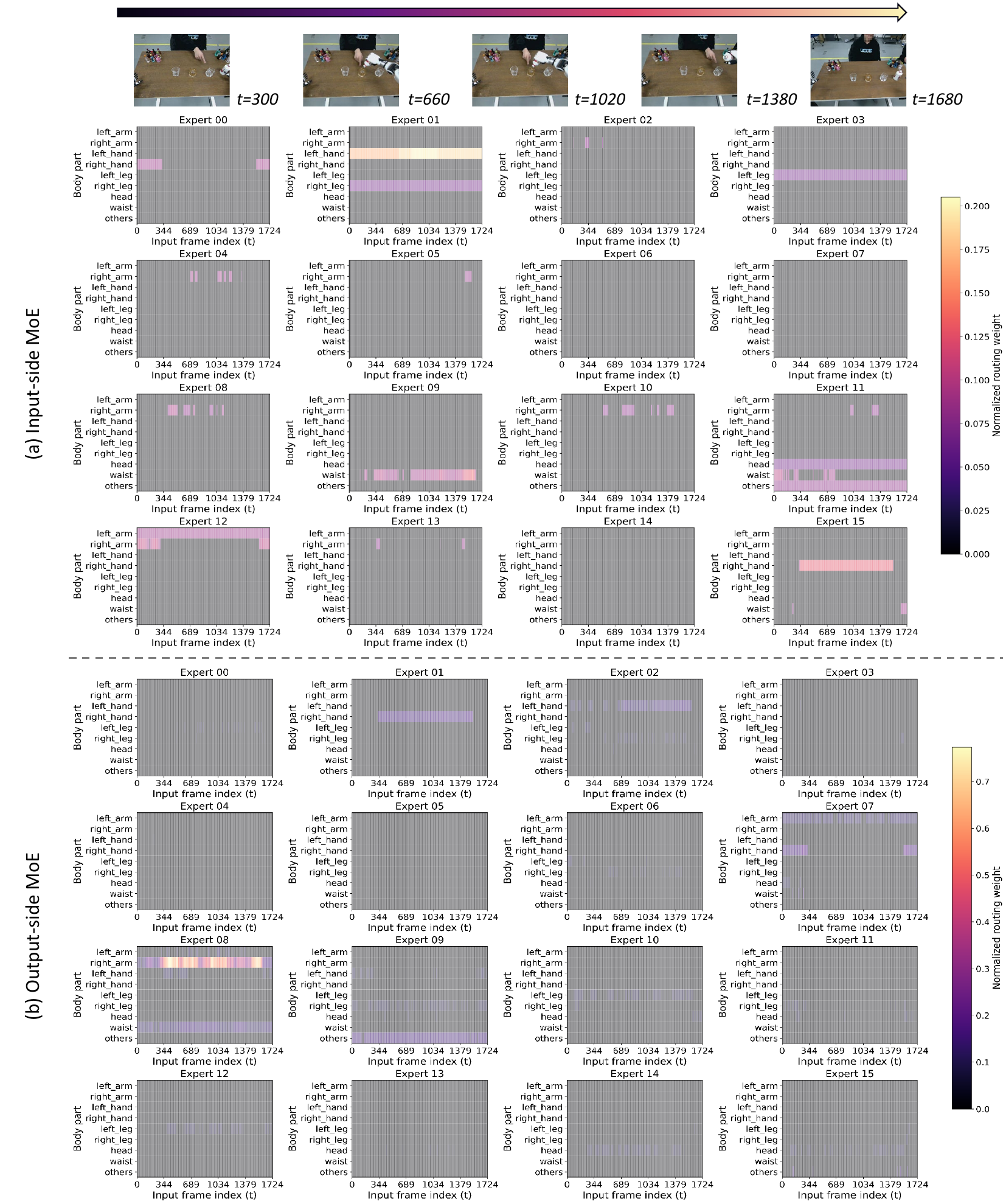}
\vskip -0.1 in
\caption{
Expert-wise routing patterns of the input-side and output-side MoE modules during a whole-body manipulation rollout.
}
\label{fig: moe_expert}
\end{figure*}

\noindent \textbf{Loss weight $\alpha$.}
The loss weight $\alpha$ controls the balance between action generation and proprioceptive prediction. Removing the state-prediction objective ($\alpha=0$) substantially degrades performance, confirming the importance of predictive proprioceptive modeling. In contrast, an excessively large weight ($\alpha=2$) also leads to severe degradation, suggesting that over-emphasizing state prediction can interfere with task-oriented action learning through the shared representation. Although $\alpha=0.5$ achieves a slightly higher SPS, $\alpha=1$ yields the highest end-to-end success rate (80.0\%), and is therefore selected as the default setting.

\begin{figure*}[t]
\centering
\includegraphics[width=1\textwidth]{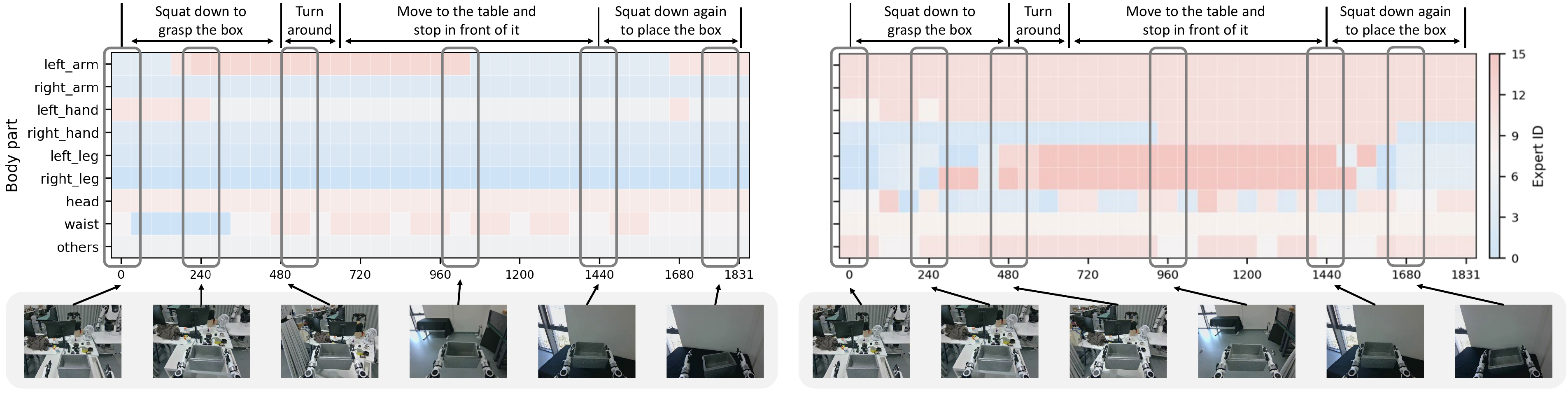}
\vskip -0.1 in
\caption{
Comparison of MoE routing patterns before and after the transformer blocks during a long-horizon box conveyance task. Left: MoE before the transformer blocks. Right: MoE after the transformer blocks. The heatmaps show the selected expert ID for each body part over time, together with representative frames and annotated subtask boundaries. Compared with the largely static routing before the transformer blocks, the routing after the transformer blocks shows clearer phase-dependent switching, suggesting stronger state-dependent control specialization.
}
\label{fig: moe_routing}
\end{figure*}

\subsubsection{MoE Routing Pattern}
Figures~\ref{fig: moe_expert} and~\ref{fig: moe_routing} analyze the routing behavior of the input-side and output-side MoE modules from complementary perspectives. Figure~\ref{fig: moe_expert} shows that the input-side MoE exhibits relatively sparse and persistent body-part preferences, indicating specialization toward different local proprioceptive components, while its activations can still vary with the evolving motion pattern. In contrast, the output-side MoE shows stronger task-dependent coordination across multiple body parts, suggesting that contextualized features after the transformer backbone enable experts to capture coordinated whole-body dynamics rather than only isolated body-part characteristics.

Figure~\ref{fig: moe_routing} further shows how these patterns evolve during long-horizon execution. Before the transformer blocks, expert assignments remain largely stable across subtask transitions, whereas after the transformer blocks, routing becomes markedly more phase-dependent, with major expert switches aligning with semantic task boundaries such as turning and locomotion. Together, these results suggest that the input-side MoE primarily captures persistent body-part-specific dynamics, while the output-side MoE further adapts expert selection to task phase and cross-part coordination.

\begin{figure*}[t]
\centering
\begin{minipage}[t]{0.48\textwidth}
    \centering
    \includegraphics[width=\linewidth]{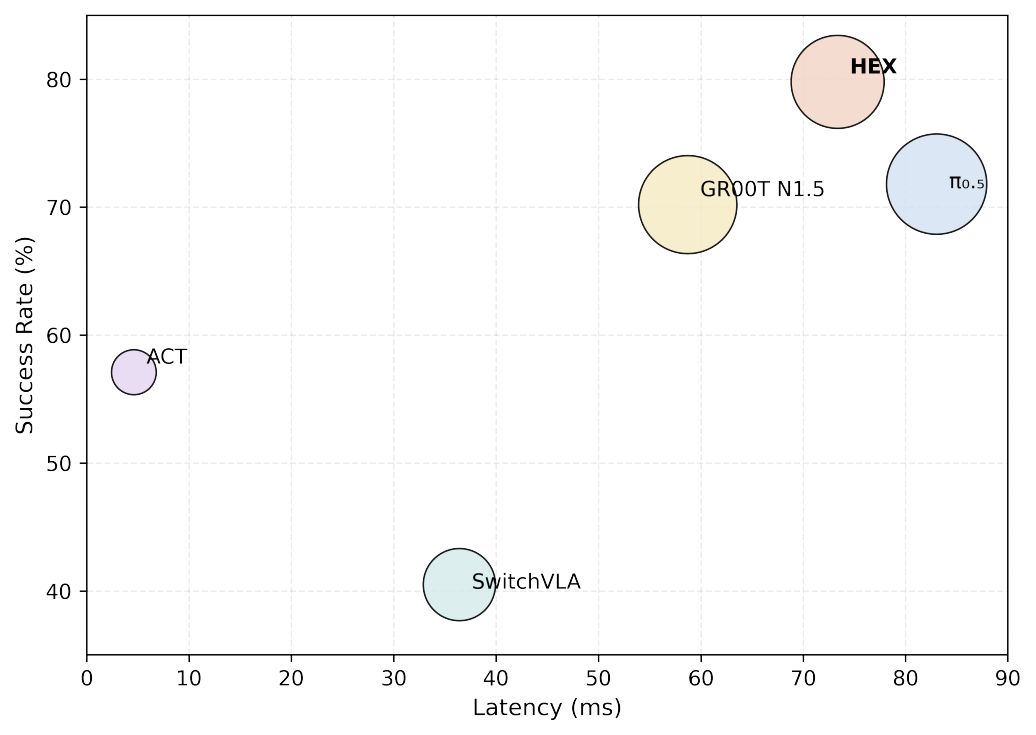}
    \vskip -0.13in
    \caption{Latency--accuracy comparison on a single NVIDIA RTX 4090 GPU, where bubble size indicates the number of model parameters.}
    \label{fig: latency}
\end{minipage}
\hfill
\begin{minipage}[t]{0.48\textwidth}
    \centering
    \includegraphics[width=\linewidth]{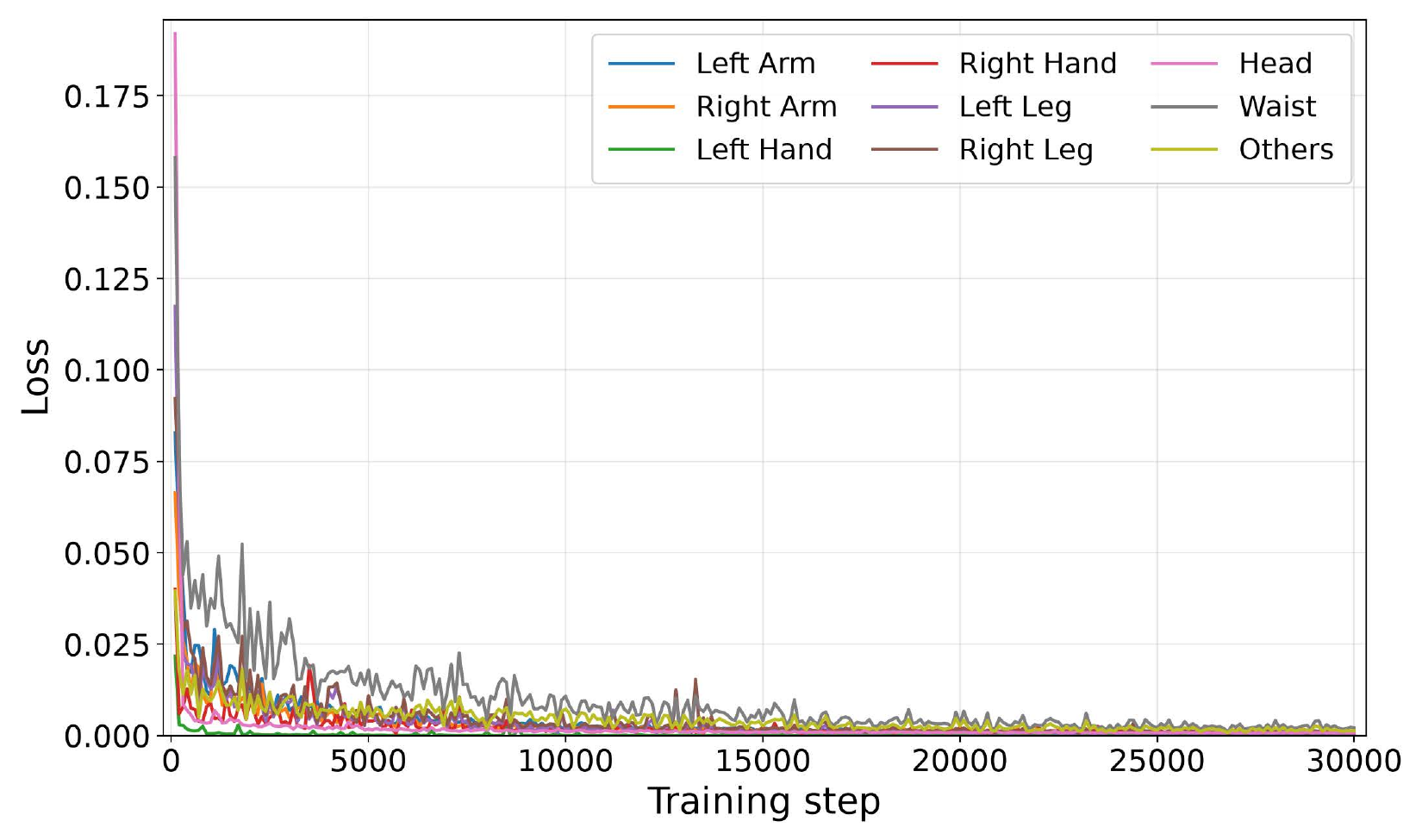}
    \vskip -0.13in
    \caption{Body-part-wise future proprioceptive prediction loss of UPP on the validation set throughout training.}
    \label{fig:upp_prediction_loss}
\end{minipage}
\end{figure*}

\noindent \textbf{Latency.}
Figure~\ref{fig: latency} compares the latency-accuracy trade-off of different methods on an RTX 4090. ACT achieves the lowest latency, but with a substantial drop in success rate. Among the large-scale baselines, GR00T N1.5 attains a relatively favorable latency-accuracy trade-off, achieving competitive performance at lower latency than both $\pi_{0.5}$ and HEX. HEX nevertheless achieves the highest success rate overall (79.8\%) with 73.34 ms latency, outperforming all baselines in task success while remaining faster than $\pi_{0.5}$. Overall, these results show that HEX provides the strongest effectiveness under a practical inference budget.

\begin{figure*}[t]
\centering
\includegraphics[width=1.0\textwidth]{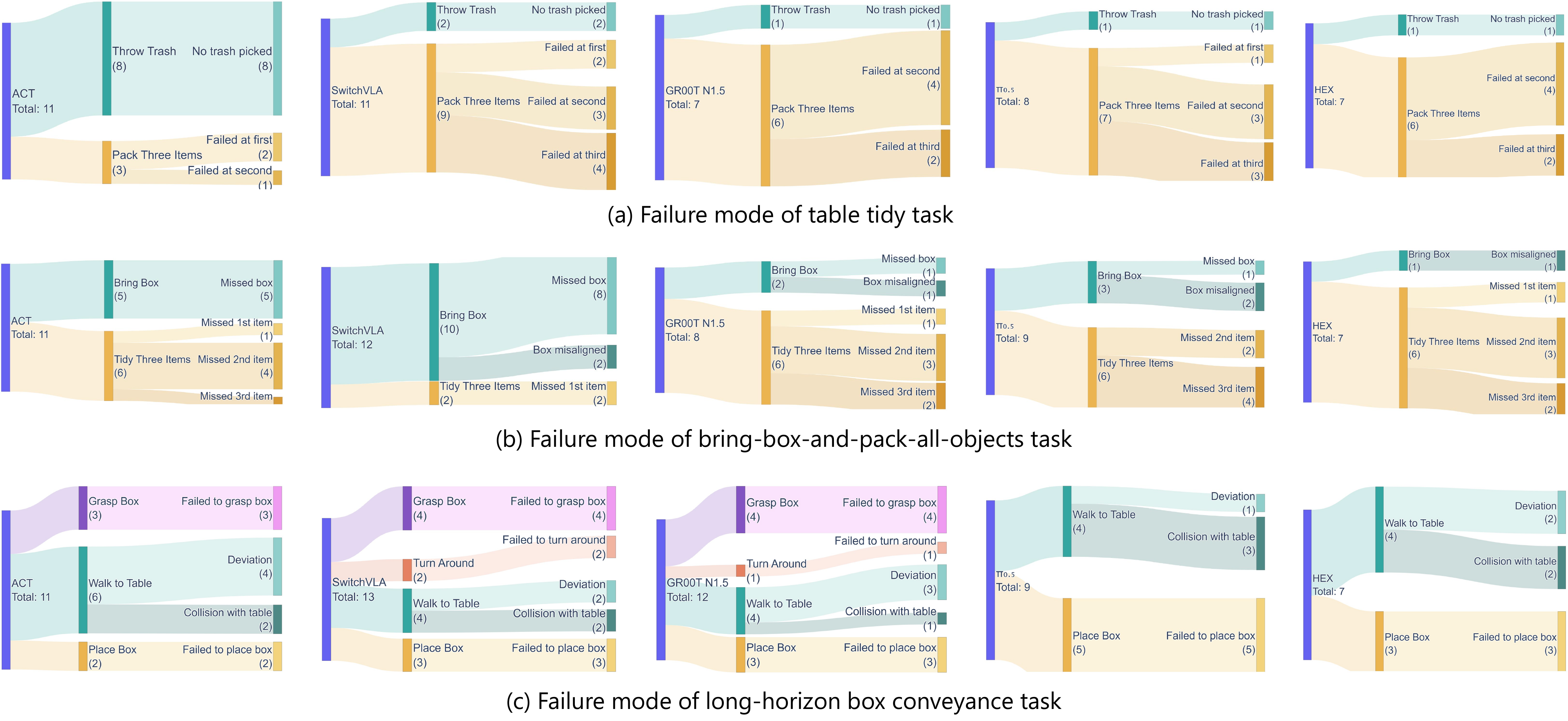}
\vskip -0.1 in
\caption{
Failure analysis across different methods and tasks. Each Sankey diagram shows how failed trials are distributed across task stages and fine-grained error types, with flow width proportional to the number of failures.
}
\label{fig: failure_analysis}
\end{figure*}

\noindent \textbf{Failure Analysis.}
Figure~\ref{fig: failure_analysis} shows that different methods fail not only at different rates, but also in different stages. In the two seen-scenario tasks, failures are relatively concentrated in a small number of key sub-stages, mainly related to object grasping, placement, and multi-object handling. In contrast, the long-horizon box conveyance task exhibits more distributed failures across grasping, turning, locomotion, and final placement, indicating that longer action chains amplify error accumulation and cross-stage dependency.
Across tasks, HEX generally yields fewer failures and a more concentrated failure distribution than the baselines. This suggests that its advantage is not only higher overall success, but also improved robustness to error propagation across sequential stages, especially in long-horizon execution.

\noindent \textbf{UPP Prediction Analysis.}
To directly examine whether UPP learns meaningful future proprioceptive dynamics, we evaluate its prediction accuracy on a held-out validation set throughout training. As shown in Fig.~\ref{fig:upp_prediction_loss}, the prediction losses of all body components steadily decrease and converge to low values, indicating that UPP can predict future proprioceptive states on unseen trajectories rather than merely fitting the training data. The body-part-wise results further reveal different prediction difficulties across components, with the waist exhibiting relatively higher residual error than most other parts, reflecting the more complex dynamics involved in whole-body coordination.

\section{Conclusion} 
\label{sec:conclusion}

We presented \textbf{HEX}, a framework for humanoid whole-body manipulation that addresses a key limitation of existing VLA-style approaches: they often do not explicitly model how different body parts interact under shared balance and posture. HEX tackles this problem through a humanoid-aligned universal state representation, predictive modeling of whole-body proprioceptive dynamics, and adaptive fusion of visual-language context with future state evolution. This leads to more coherent and stable whole-body action generation.
Extensive experiments on real-world humanoid manipulation tasks show that HEX achieves superior performance over strong baselines, particularly in fast-reaction and long-horizon settings where coordinated whole-body behavior is essential. Overall, our results highlight the importance of explicitly modeling structured body-part interaction for general and scalable humanoid manipulation.

{
\small
\bibliographystyle{plain}
\bibliography{main}

@article{nakanishi2004learning,
  title={Learning from demonstration and adaptation of biped locomotion},
  author={Nakanishi, Jun and Morimoto, Jun and Endo, Gen and Cheng, Gordon and Schaal, Stefan and Kawato, Mitsuo},
  journal={Robotics and autonomous systems},
  volume={47},
  number={2-3},
  pages={79--91},
  year={2004},
  publisher={Elsevier}
}

@article{radosavovic2024real,
  title={Real-world humanoid locomotion with reinforcement learning},
  author={Radosavovic, Ilija and Xiao, Tete and Zhang, Bike and Darrell, Trevor and Malik, Jitendra and Sreenath, Koushil},
  journal={Science Robotics},
  volume={9},
  number={89},
  pages={eadi9579},
  year={2024},
  publisher={American Association for the Advancement of Science}
}

@inproceedings{li2026language,
  title={From language to locomotion: Retargeting-free humanoid control via motion latent guidance},
  author={Li, Zhe and Wei, Yangyang and Zhu, Boan and Peng, Yibo and Huang, Tao and Wang, Pengwei and Wang, Zhongyuan and Chi, Cheng and Xu, Chang and Zhang, Shanghang},
  booktitle={International Conference on Learning Representations},
  volume={2026},
  pages={129292--129319},
  year={2026}
}

@inproceedings{li2026robomirror,
  title={Robomirror: Understand before you imitate for video to humanoid locomotion},
  author={Li, Zhe and Zhu, Boan and Wei, Yangyang and Bai, Shuanghao and Ji, Yuheng and Peng, Yibo and Huang, Tao and Wang, Pengwei and Wang, Zhongyuan and Chan, S-H Gary and others},
  booktitle={European Conference on Computer Vision},
  pages={482--499},
  year={2026},
  organization={Springer}
}

@inproceedings{li2025okami,
  title={OKAMI: Teaching Humanoid Robots Manipulation Skills through Single Video Imitation},
  author={Li, Jinhan and Zhu, Yifeng and Xie, Yuqi and Jiang, Zhenyu and Seo, Mingyo and Pavlakos, Georgios and Zhu, Yuke},
  booktitle={Conference on Robot Learning},
  pages={299--317},
  year={2025},
  organization={PMLR}
}

@article{xie2023hierarchical,
  title={Hierarchical planning and control for box loco-manipulation},
  author={Xie, Zhaoming and Tseng, Jonathan and Starke, Sebastian and Van De Panne, Michiel and Liu, C Karen},
  journal={Proceedings of the ACM on Computer Graphics and Interactive Techniques},
  volume={6},
  number={3},
  pages={1--18},
  year={2023},
  publisher={ACM New York, NY, USA}
}

@inproceedings{chen2025hand,
  title={Hand-Eye Autonomous Delivery: Learning Humanoid Navigation, Locomotion and Reaching},
  author={Chen, Sirui and Ye, Yufei and Cao, Zi-ang and Xu, Pei and Lew, Jennifer and Liu, Karen},
  booktitle={Conference on Robot Learning},
  pages={4058--4073},
  year={2025},
  organization={PMLR}
}

@article{peng2018deepmimic,
  title={Deepmimic: Example-guided deep reinforcement learning of physics-based character skills},
  author={Peng, Xue Bin and Abbeel, Pieter and Levine, Sergey and Van de Panne, Michiel},
  journal={ACM Transactions On Graphics (TOG)},
  volume={37},
  number={4},
  pages={1--14},
  year={2018},
  publisher={ACM New York, NY, USA}
}

@article{peng2021amp,
  title={Amp: Adversarial motion priors for stylized physics-based character control},
  author={Peng, Xue Bin and Ma, Ze and Abbeel, Pieter and Levine, Sergey and Kanazawa, Angjoo},
  journal={ACM Transactions on Graphics (ToG)},
  volume={40},
  number={4},
  pages={1--20},
  year={2021},
  publisher={ACM New York, NY, USA}
}

@inproceedings{hu2025slac,
  title={SLAC: Simulation-Pretrained Latent Action Space for Whole-Body Real-World RL},
  author={Hu, Jiaheng and Stone, Peter and Mart{\'\i}n-Mart{\'\i}n, Roberto},
  booktitle={Conference on Robot Learning},
  pages={2966--2982},
  year={2025},
  organization={PMLR}
}

@inproceedings{zhang2026falcon,
  title={FALCON: Learning Force-Adaptive Humanoid Loco-Manipulation},
  author={Zhang, Yuanhang and Yuan, Yifu and Gurunath, Prajwal and Gupta, Ishita and Omidshafiei, Shayegan and Agha-mohammadi, Ali-akbar and Vazquez-Chanlatte, Marcell and Pedersen, Liam and He, Tairan and Shi, Guanya},
  booktitle={Proceedings of the 8th Annual Learning for Dynamics and Control Conference},
  pages={265--281},
  year={2026},
  volume={331},
  publisher={PMLR}
}

@inproceedings{xie2026kungfubot,
  title={Kungfubot: Physics-based humanoid whole-body control for learning highly-dynamic skills},
  author={Xie, Weiji and Han, Jinrui and Zheng, Jiakun and Li, Huanyu and Liu, Xinzhe and Shi, Jiyuan and Zhang, Weinan and Bai, Chenjia and Li, Xuelong},
  booktitle={Advances in Neural Information Processing Systems},
  volume={38},
  pages={62406--62433},
  year={2026}
}

@article{luo2026sonic,
  title={Sonic: Supersizing motion tracking for natural humanoid whole-body control},
  author={Luo, Zhengyi and Yuan, Ye and Wang, Tingwu and Li, Chenran and Casta{\~n}eda, Fernando and Chen, Sirui and Cao, Zi-Ang and Li, Jiefeng and Minor, David and Ben, Qingwei and others},
  journal={Science Robotics},
  volume={11},
  number={117},
  pages={eaed4592},
  year={2026},
  publisher={American Association for the Advancement of Science}
}

@inproceedings{he2026viral,
  title={Viral: Visual sim-to-real at scale for humanoid loco-manipulation},
  author={He, Tairan and Wang, Zi and Xue, Haoru and Ben, Qingwei and Luo, Zhengyi and Xiao, Wenli and Yuan, Ye and Da, Xingye and Casta{\~n}eda, Fernando and Sastry, Shankar and others},
  booktitle={Proceedings of the IEEE/CVF Conference on Computer Vision and Pattern Recognition},
  pages={13430--13441},
  year={2026}
}

@article{liao2026beyondmimic,
  title={Beyondmimic: From motion tracking to versatile humanoid control via guided diffusion},
  author={Liao, Qiayuan and Truong, Takara E and Huang, Xiaoyu and Gao, Yuman and Tevet, Guy and Sreenath, Koushil and Liu, C Karen},
  journal={Science Robotics},
  volume={11},
  number={117},
  pages={eadx8924},
  year={2026},
  publisher={American Association for the Advancement of Science}
}

@inproceedings{fu2025humanplus,
  title={HumanPlus: Humanoid Shadowing and Imitation from Humans},
  author={Fu, Zipeng and Zhao, Qingqing and Wu, Qi and Wetzstein, Gordon and Finn, Chelsea},
  booktitle={Conference on Robot Learning},
  pages={2828--2844},
  year={2025},
  organization={PMLR}
}

@inproceedings{he2025omnih2o,
  title={OmniH2O: Universal and Dexterous Human-to-Humanoid Whole-Body Teleoperation and Learning},
  author={He, Tairan and Luo, Zhengyi and He, Xialin and Xiao, Wenli and Zhang, Chong and Zhang, Weinan and Kitani, Kris M and Liu, Changliu and Shi, Guanya},
  booktitle={Conference on Robot Learning},
  pages={1516--1540},
  year={2025},
  organization={PMLR}
}

@article{chen2025gmt,
  title={Gmt: General motion tracking for humanoid whole-body control},
  author={Chen, Zixuan and Ji, Mazeyu and Cheng, Xuxin and Peng, Xuanbin and Peng, Xue Bin and Wang, Xiaolong},
  journal={arXiv preprint arXiv:2506.14770},
  year={2025}
}

@inproceedings{ze2025generalizable,
  title={Generalizable humanoid manipulation with 3d diffusion policies},
  author={Ze, Yanjie and Chen, Zixuan and Wang, Wenhao and Chen, Tianyi and He, Xialin and Yuan, Ying and Peng, Xue Bin and Wu, Jiajun},
  booktitle={2025 IEEE/RSJ International Conference on Intelligent Robots and Systems (IROS)},
  pages={2873--2880},
  year={2025},
  organization={IEEE}
}

@inproceedings{lu2025mobile,
  title={Mobile-television: Predictive motion priors for humanoid whole-body control},
  author={Lu, Chenhao and Cheng, Xuxin and Li, Jialong and Yang, Shiqi and Ji, Mazeyu and Yuan, Chengjing and Yang, Ge and Yi, Sha and Wang, Xiaolong},
  booktitle={2025 IEEE International Conference on Robotics and Automation (ICRA)},
  pages={5364--5371},
  year={2025},
  organization={IEEE}
}

@inproceedings{ze2025twist,
  title={TWIST: Teleoperated Whole-Body Imitation System},
  author={Ze, Yanjie and Chen, Zixuan and Araujo, Joao Pedro and Cao, Zi-ang and Peng, Xue Bin and Wu, Jiajun and Liu, Karen},
  booktitle={Conference on Robot Learning},
  pages={2143--2154},
  year={2025},
  organization={PMLR}
}

@article{ding2025humanoid,
  title={Humanoid-vla: Towards universal humanoid control with visual integration},
  author={Ding, Pengxiang and Ma, Jianfei and Tong, Xinyang and Zou, Binghong and Luo, Xinxin and Fan, Yiguo and Wang, Ting and Lu, Hongchao and Mo, Panzhong and Liu, Jinxin and others},
  journal={arXiv preprint arXiv:2502.14795},
  year={2025}
}

@article{xue2025leverb,
  title={Leverb: Humanoid whole-body control with latent vision-language instruction},
  author={Xue, Haoru and Huang, Xiaoyu and Niu, Dantong and Liao, Qiayuan and Kragerud, Thomas and Gravdahl, Jan Tommy and Peng, Xue Bin and Shi, Guanya and Darrell, Trevor and Sreenath, Koushil and others},
  journal={arXiv preprint arXiv:2506.13751},
  year={2025}
}

@inproceedings{liu2026trajbooster,
  title={Trajbooster: Boosting humanoid whole-body manipulation via trajectory-centric learning},
  author={Liu, Jiacheng and Ding, Pengxiang and Zhou, Qihang and Wu, Yuxuan and Huang, Da and Peng, Zimian and Xiao, Wei and Zhang, Weinan and Yang, Lixin and Lu, Cewu and others},
  booktitle={2026 IEEE International Conference on Robotics and Automation (ICRA)},
  pages={17257--17264},
  year={2026},
  organization={IEEE}
}

@inproceedings{jiang2026wholebodyvla,
  title={Wholebodyvla: Towards unified latent vla for whole-body loco-manipulation control},
  author={Jiang, Haoran and Chen, Jin and Bu, Qingwen and Chen, Li and Shi, Modi and Zhang, Yanjie and Li, Delong and Suo, Chuanzhe and Li, Hongyang and others},
  booktitle={International Conference on Learning Representations},
  volume={2026},
  pages={157438--157461},
  year={2026}
}

@article{zhao2026halomi,
  title={HALOMI: Learning Humanoid Loco-Manipulation with Active Perception from Human Demonstrations},
  author={Zhao, Zehui and Zhao, Yuxuan and Zhang, Gaojing and Liu, Chenxi and Zheng, Maolin and Lian, Wenzhao},
  journal={arXiv preprint arXiv:2606.18772},
  year={2026}
}

@article{hu2026openhlm,
  title={OpenHLM: An Empirical Recipe for Whole-Body Humanoid Loco-Manipulation},
  author={Hu, Yingdong and Zhu, Haodong and Zheng, Boyuan and Hu, Yihang and Zhang, Tong and Chen, Zunhao and Zhao, Junming and Nai, Ruiqian and Gao, Yang},
  journal={arXiv preprint arXiv:2606.22174},
  year={2026}
}

@article{wang2026vlk,
  title={VLK: Learning Humanoid Loco-Manipulation from Synthetic Interactions in Reconstructed Scenes},
  author={Wang, Yen-Jen and Li, Jiaman and Chen, Sirui and Truong, Takara E and Xu, Pei and Abbeel, Pieter and Duan, Rocky and Sreenath, Koushil and Kanazawa, Angjoo and Sferrazza, Carmelo and others},
  journal={arXiv preprint arXiv:2606.30645},
  year={2026}
}

@inproceedings{gu2026haf,
  title={HAF: Adapting Generalist VLAs to Humanoid Whole-Body Loco-manipulation via Hierarchical Action Flow and Spectral Latent RL},
  author={Gu, Langzhe and Hou, Chengkai and Li, Meng and Wang, Xinhua and Liu, Jiaming and Lv, Xinyuan and Zhang, Bowei and Bai, Shuanghao and Li, Guangrun and He, Jingyang and others},
  booktitle={Conference on Robot Learning},
  year={2026}
}

@article{zheng2026motionwam,
  title={MotionWAM: Towards Foundation World Action Models for Real-Time Humanoid Loco-Manipulation},
  author={Zheng, Jia and Ma, Teli and Fan, Yudong and Wang, Zifan and Yang, Shuo and Liang, Junwei},
  journal={arXiv preprint arXiv:2606.09215},
  year={2026}
}

@article{li2026omega,
  title={$\omega$-0: A Latent Predictive World Action Model for Concurrent Humanoid Loco-Manipulation},
  author={Li, Zhe and Zhang, Zhenzhe and Wei, Yangyang and Zhang, Wenjie and Yuan, Xichen and Zhi, Peiyuan and Li, Gen and Guo, Xinying and Gao, Fengjie and Yang, Jianfei and others},
  journal={arXiv preprint arXiv:2608.06375},
  year={2026}
}

@article{li2026wholebodywam,
  title={WholeBodyWAM: Generalizing Pre-trained World-Action Priors to Humanoid Loco-Manipulation via WBC-Grounded Coordination},
  author={Li, Zhuo and Yao, Yiming and Tan, Jim and Jing, Mengjie and Dong, Zhipeng and Chen, Fei},
  journal={arXiv preprint arXiv:2609.16644},
  year={2026}
}

@article{zhang2026wholebodywam,
  title={WholeBodyWAM: Learning Whole-Body World Action Models with Scalable Motion Priors},
  author={Zhang, Bowei and Zhang, Qiyao and Bai, Shuanghao and Wang, Xinhua and Li, Meng and Wang, Yilei and Zhang, Leiwang and Tang, Jian and Zhou, Lu and Sun, Lei and others},
  journal={arXiv preprint arXiv:2609.18197},
  year={2026}
}

@inproceedings{mao2025universal,
  title={Universal humanoid robot pose learning from internet human videos},
  author={Mao, Jiageng and Zhao, Siheng and Song, Siqi and Hong, Chuye and Shi, Tianheng and Ye, Junjie and Zhang, Mingtong and Geng, Haoran and Malik, Jitendra and Guizilini, Vitor and others},
  booktitle={2025 IEEE-RAS 24th International Conference on Humanoid Robots (Humanoids)},
  pages={1--8},
  year={2025},
  organization={IEEE}
}

@inproceedings{qiu2025humanoid,
  title={Humanoid Policy Human Policy},
  author={Qiu, Ri-Zhao and Yang, Shiqi and Cheng, Xuxin and Chawla, Chaitanya and Li, Jialong and He, Tairan and Yan, Ge and Yoon, David J and Hoque, Ryan and Paulsen, Lars and others},
  booktitle={Conference on Robot Learning},
  pages={2888--2906},
  year={2025},
  organization={PMLR}
}

@inproceedings{punamiya2025egobridge,
  title={EgoBridge: Domain Adaptation for Generalizable Imitation from Egocentric Human Data},
  author={Punamiya, Ryan and Patel, Dhruv and Aphiwetsa, Patcharapong and Kuppili, Pranav and Zhu, Lawrence and Kareer, Simar and Hoffman, Judy and Xu, Danfei},
  booktitle={Advances in Neural Information Processing Systems},
  volume={38},
  pages={47092--47122},
  year={2025}
}

@article{weng2025hdmi,
  title={Hdmi: Learning interactive humanoid whole-body control from human videos},
  author={Weng, Haoyang and Li, Yitang and Sobanbabu, Nikhil and Wang, Zihan and Luo, Zhengyi and He, Tairan and Ramanan, Deva and Shi, Guanya},
  journal={arXiv preprint arXiv:2509.16757},
  year={2025}
}

@article{yang2025egovla,
  title={Egovla: Learning vision-language-action models from egocentric human videos},
  author={Yang, Ruihan and Yu, Qinxi and Wu, Yecheng and Yan, Rui and Li, Borui and Cheng, An-Chieh and Zou, Xueyan and Fang, Yunhao and Cheng, Xuxin and Qiu, Ri-Zhao and others},
  journal={arXiv preprint arXiv:2507.12440},
  year={2025}
}

@inproceedings{shi2026egohumanoid,
  title={EgoHumanoid: Unlocking In-the-Wild Loco-Manipulation with Robot-Free Egocentric Demonstration},
  author={Shi, Modi and Peng, Shijia and Chen, Jin and Jiang, Haoran and Li, Yinghui and Huang, Di and Luo, Ping and Li, Hongyang and Chen, Li},
  booktitle={Robotics: Science and Systems},
  year={2026}
}

@article{yang2026zerowbc,
  title={ZeroWBC: Learning Natural Visuomotor Humanoid Control Directly from Human Egocentric Video},
  author={Yang, Haoran and Bao, Jiacheng and Xin, Yucheng and Song, Haoming and Tian, Yuyang and Zhao, Bin and Wang, Dong and Li, Xuelong},
  journal={arXiv preprint arXiv:2603.09170},
  year={2026}
}

@inproceedings{wei2026psi0,
  title={{$\Psi_0$}: An Open Foundation Model Towards Universal Humanoid Loco-Manipulation},
  author={Wei, Songlin and Jing, Hongyi and Li, Boqian and Zhao, Zhenyu and Mao, Jiageng and Ni, Zhenhao and He, Sicheng and Liu, Jie and Liu, Xiawei and Kang, Kaidi and others},
  booktitle={Robotics: Science and Systems},
  year={2026}
}

@inproceedings{lin2026h,
  title={H-zero: Cross-humanoid locomotion pretraining enables few-shot novel embodiment transfer},
  author={Lin, Yunfeng and Liu, Minghuan and Xue, Yufei and Zhou, Ming and Yu, Yong and Pang, Jiangmiao and Zhang, Weinan},
  booktitle={2026 IEEE International Conference on Robotics and Automation (ICRA)},
  pages={9306--9313},
  year={2026},
  organization={IEEE}
}

@inproceedings{xue2026scalable,
  title={Scalable and General Whole-Body Control for Cross-Humanoid Locomotion},
  author={Xue, Yufei and Lin, Yunfeng and Dong, Wentao and Tang, Yang and Wang, Jingbo and Pang, Jiangmiao and Zhou, Ming and Liu, Minghuan and Zhang, Weinan},
  booktitle={Proceedings of the 43rd International Conference on Machine Learning},
  pages={143236--143254},
  year={2026},
  volume={306},
  publisher={PMLR}
}

@inproceedings{peng2026embodiment,
  title={Embodiment-Aware Generalist Specialist Distillation for Unified Humanoid Whole-Body Control},
  author={Peng, Quanquan and Lin, Yunfeng and Xue, Yufei and Pang, Jiangmiao and Zhang, Weinan},
  booktitle={2026 IEEE International Conference on Robotics and Automation (ICRA)},
  pages={16996--17003},
  year={2026}
}

@article{bjorck2025gr00t,
  title={Gr00t n1: An open foundation model for generalist humanoid robots},
  author={Bjorck, Johan and Casta{\~n}eda, Fernando and Cherniadev, Nikita and Da, Xingye and Ding, Runyu and Fan, Linxi and Fang, Yu and Fox, Dieter and Hu, Fengyuan and Huang, Spencer and others},
  journal={arXiv preprint arXiv:2503.14734},
  year={2025}
}

@article{luo2026being,
  title={Being-H0.5: Scaling Human-Centric Robot Learning for Cross-Embodiment Generalization},
  author={Luo, Hao and Wang, Ye and Zhang, Wanpeng and Zheng, Sipeng and Xi, Ziheng and Xu, Chaoyi and Xu, Haiweng and Yuan, Haoqi and Zhang, Chi and Wang, Yiqing and others},
  journal={arXiv preprint arXiv:2601.12993},
  year={2026}
}

@inproceedings{wang2024scaling,
  title={Scaling proprioceptive-visual learning with heterogeneous pre-trained transformers},
  author={Wang, Lirui and Chen, Xinlei and Zhao, Jialiang and He, Kaiming},
  booktitle={Advances in neural information processing systems},
  volume={37},
  pages={124420--124450},
  year={2024}
}

@inproceedings{yang2024pushing,
  title={Pushing the limits of cross-embodiment learning for manipulation and navigation},
  author={Yang, Jonathan and Glossop, Catherine and Bhorkar, Arjun and Shah, Dhruv and Vuong, Quan and Finn, Chelsea and Sadigh, Dorsa and Levine, Sergey},
  booktitle={Robotics: Science and Systems}, 
  year={2024}
}

@inproceedings{doshi2025scaling,
  title={Scaling Cross-Embodied Learning: One Policy for Manipulation, Navigation, Locomotion and Aviation},
  author={Doshi, Ria and Walke, Homer Rich and Mees, Oier and Dasari, Sudeep and Levine, Sergey},
  booktitle={Conference on Robot Learning},
  pages={496--512},
  year={2025},
  organization={PMLR}
}

@inproceedings{kim2025openvla,
  title={OpenVLA: An Open-Source Vision-Language-Action Model},
  author={Kim, Moo Jin and Pertsch, Karl and Karamcheti, Siddharth and Xiao, Ted and Balakrishna, Ashwin and Nair, Suraj and Rafailov, Rafael and Foster, Ethan P and Sanketi, Pannag R and Vuong, Quan and others},
  booktitle={Conference on Robot Learning},
  pages={2679--2713},
  year={2025},
  organization={PMLR}
}

@inproceedings{intelligence2025pi05,
  title={{$\pi_{0.5}$}: A Vision-Language-Action Model with Open-World Generalization},
  author={Intelligence, Physical and Black, Kevin and Brown, Noah and Darpinian, James and Dhabalia, Karan and Driess, Danny and Esmail, Adnan and Equi, Michael and Finn, Chelsea and Fusai, Niccolo and others},
  booktitle={Conference on Robot Learning},
  pages={17--40},
  year={2025},
  volume={305},
  publisher={PMLR}
}

@article{cui2025openhelix,
  title={Openhelix: A short survey, empirical analysis, and open-source dual-system vla model for robotic manipulation},
  author={Cui, Can and Ding, Pengxiang and Song, Wenxuan and Bai, Shuanghao and Tong, Xinyang and Ge, Zirui and Suo, Runze and Zhou, Wanqi and Liu, Yang and Jia, Bofang and others},
  journal={arXiv preprint arXiv:2505.03912},
  year={2025}
}

@inproceedings{zhao2023learning,
  title={Learning Fine-Grained Bimanual Manipulation with Low-Cost Hardware},
  author={Zhao, Tony and Kumar, Vikash and Levine, Sergey and Finn, Chelsea},
  booktitle={Robotics: Science and Systems},
  year={2023}
}

@article{li2025switchvla,
  title={Switchvla: Execution-aware task switching for vision-language-action models},
  author={Li, Meng and Zhao, Zhen and Che, Zhengping and Liao, Fei and Wu, Kun and Xu, Zhiyuan and Ren, Pei and Jin, Zhao and Liu, Ning and Tang, Jian},
  journal={arXiv preprint arXiv:2506.03574},
  year={2025}
}

@inproceedings{zhao2026humanoid,
  title={Humanoid everyday: A comprehensive robotic dataset for open-world humanoid manipulation},
  author={Zhao, Zhenyu and Jing, Hongyi and Liu, Xiawei and Mao, Jiageng and Jha, Abha and Yang, Hanwen and Xue, Rong and Zakharov, Sergey and Guizilini, Vitor and Wang, Yue},
  booktitle={2026 IEEE International Conference on Robotics and Automation (ICRA)},
  pages={6415--6422},
  year={2026},
  organization={IEEE}
}

@inproceedings{bu2025agibot,
  title={Agibot world colosseo: A large-scale manipulation platform for scalable and intelligent embodied systems},
  author={Bu, Qingwen and Cai, Jisong and Chen, Li and Cui, Xiuqi and Ding, Yan and Feng, Siyuan and Gao, Shenyuan and He, Xindong and Hu, Xuan and Huang, Xu and others},
  booktitle={2025 IEEE/RSJ International Conference on Intelligent Robots and Systems (IROS)},
  pages={3549-3556},
  year={2025}
}

@article{wu2025robocoin,
  title={RoboCOIN: An Open-Sourced Bimanual Robotic Data COllection for INtegrated Manipulation},
  author={Wu, Shihan and Liu, Xuecheng and Xie, Shaoxuan and Wang, Pengwei and Li, Xinghang and Yang, Bowen and Li, Zhe and Zhu, Kai and Wu, Hongyu and Liu, Yiheng and others},
  journal={arXiv preprint arXiv:2511.17441},
  year={2025}
}

@article{bai2025qwen3,
  title={Qwen3-vl technical report},
  author={Bai, Shuai and Cai, Yuxuan and Chen, Ruizhe and Chen, Keqin and Chen, Xionghui and Cheng, Zesen and Deng, Lianghao and Ding, Wei and Gao, Chang and Ge, Chunjiang and others},
  journal={arXiv preprint arXiv:2511.21631},
  year={2025}
}

@article{du2025himoe,
  title={HiMoE-VLA: Hierarchical Mixture-of-Experts for Generalist Vision-Language-Action Policies},
  author={Du, Zhiying and Liu, Bei and Liang, Yaobo and Shen, Yichao and Cao, Haidong and Zheng, Xiangyu and Feng, Zhiyuan and Wu, Zuxuan and Yang, Jiaolong and Jiang, Yu-Gang},
  journal={arXiv preprint arXiv:2512.05693},
  year={2025}
}

@inproceedings{xu2025hacts,
  title={Hacts: a human-as-copilot teleoperation system for robot learning},
  author={Xu, Zhiyuan and Zhao, Yinuo and Wu, Kun and Liu, Ning and Ji, Junjie and Che, Zhengping and Liu, Chi Harold and Tang, Jian},
  booktitle={2025 IEEE/RSJ International Conference on Intelligent Robots and Systems (IROS)},
  pages={15475--15481},
  year={2025},
  organization={IEEE}
}

@inproceedings{wu2025robomind,
  title={Robomind: Benchmark on multi-embodiment intelligence normative data for robot manipulation},
  author={Wu, Kun and Hou, Chengkai and Liu, Jiaming and Che, Zhengping and Ju, Xiaozhu and Yang, Zhuqin and Li, Meng and Zhao, Yinuo and Xu, Zhiyuan and Yang, Guang and others},
  booktitle={Robotics: Science and Systems}, 
  year={2025},
}

@article{hou2025robomind,
  title={Robomind 2.0: A multimodal, bimanual mobile manipulation dataset for generalizable embodied intelligence},
  author={Hou, Chengkai and Wu, Kun and Liu, Jiaming and Che, Zhengping and Wu, Di and Liao, Fei and Li, Guangrun and He, Jingyang and Feng, Qiuxuan and Jin, Zhao and others},
  journal={arXiv preprint arXiv:2512.24653},
  year={2025}
}

@article{bai2025towards,
  title={Towards a unified understanding of robot manipulation: A comprehensive survey},
  author={Bai, Shuanghao and Song, Wenxuan and Chen, Jiayi and Ji, Yuheng and Zhong, Zhide and Yang, Jin and Zhao, Han and Zhou, Wanqi and Zhao, Wei and Li, Zhe and others},
  journal={arXiv preprint arXiv:2510.10903},
  year={2025}
}

@inproceedings{fan2025long,
  title={Long-VLA: Unleashing Long-Horizon Capability of Vision Language Action Model for Robot Manipulation},
  author={Fan, Yiguo and Bai, Shuanghao and Tong, Xinyang and Ding, Pengxiang and Zhu, Yuyang and Lu, Hongchao and Dai, Fengqi and Zhao, Wei and Liu, Yang and Huang, Siteng and others},
  booktitle={Conference on Robot Learning},
  pages={2018--2037},
  year={2025},
  organization={PMLR}
}

@article{bai2026embodied,
  title={Embodied robot manipulation in the era of foundation models: Planning and learning perspectives},
  author={Bai, Shuanghao and Song, Wenxuan and Chen, Jiayi and Ji, Yuheng and Zhong, Zhide and Yang, Jin and Zhao, Han and Zhou, Wanqi and Li, Zhe and Ding, Pengxiang and others},
  journal={IEEE Transactions on Robotics},
  year={2026},
  publisher={IEEE}
}

@inproceedings{bai2026latent,
  title={Latent Reasoning {VLA}: Latent Thinking and Prediction for Vision-Language-Action Models},
  author={Bai, Shuanghao and Lyu, Jing and Zhou, Wanqi and Li, Zhe and Wang, Dakai and Xing, Lei and Zhao, Xiaoguang and Wang, Pengwei and Wang, Zhongyuan and Chi, Cheng and Chen, Badong and Zhang, Shanghang},
  booktitle={Proceedings of the 43rd International Conference on Machine Learning},
  pages={5409--5426},
  year={2026},
  volume={306},
  publisher={PMLR}
}

@inproceedings{bu2025univla,
  title={Univla: Learning to act anywhere with task-centric latent actions},
  author={Bu, Qingwen and Yang, Yanting and Cai, Jisong and Gao, Shenyuan and Ren, Guanghui and Yao, Maoqing and Luo, Ping and Li, Hongyang},
  booktitle={Robotics: Science and Systems},
  year={2025}
}

@inproceedings{song2026reconvla,
  title={Reconvla: Reconstructive vision-language-action model as effective robot perceiver},
  author={Song, Wenxuan and Zhou, Ziyang and Zhao, Han and Chen, Jiayi and Ding, Pengxiang and Yan, Haodong and Huang, Yuxin and Tang, Feilong and Wang, Donglin and Li, Haoang},
  booktitle={Proceedings of the AAAI Conference on Artificial Intelligence},
  volume={40},
  number={22},
  pages={18549--18557},
  year={2026}
}

@inproceedings{wang2026vlaadapter,
  title={Vla-adapter: An effective paradigm for tiny-scale vision-language-action model},
  author={Wang, Yihao and Ding, Pengxiang and Li, Lingxiao and Cui, Can and Ge, Zirui and Tong, Xinyang and Song, Wenxuan and Zhao, Han and Zhao, Wei and Hou, Pengxu and others},
  booktitle={Proceedings of the AAAI conference on artificial intelligence},
  volume={40},
  number={22},
  pages={18638--18646},
  year={2026}
}
}

\end{document}